\PassOptionsToPackage{table}{xcolor}
\documentclass{article}
\usepackage{iclr2027_conference,times}

\usepackage{amsmath,amsfonts,bm}

\def\eqref#1{equation~\ref{#1}}
\def\1{\bm{1}}

\DeclareMathAlphabet{\mathsfit}{\encodingdefault}{\sfdefault}{m}{sl}
\SetMathAlphabet{\mathsfit}{bold}{\encodingdefault}{\sfdefault}{bx}{n}

\usepackage{hyperref}
\usepackage{url}
\definecolor{darkblue}{rgb}{0, 0.08, 0.45}
\hypersetup{colorlinks=true,linkcolor=darkblue,citecolor=darkblue,urlcolor=darkblue}

\ifdefined\pdfgentounicode
  \usepackage{cmap}
  \input{glyphtounicode}
\fi

\usepackage{latexsym}
\usepackage[T1]{fontenc}
\usepackage[utf8]{inputenc}
\usepackage{microtype}
\usepackage{amsmath}
\usepackage{amssymb}
\usepackage{eurosym}
\usepackage{graphicx}
\usepackage{tikz}
\usepackage{booktabs}
\usepackage{colortbl}
\usepackage{multirow}

\usetikzlibrary{arrows.meta,positioning,calc,backgrounds,fit,shapes.geometric}

\newcommand{\ssub}[1]{{\scriptsize\textcolor{gray}{(#1)}}}
\newcommand{\std}[1]{{\scriptsize\textcolor{gray}{$\pm$#1}}}

\graphicspath{{figures/}{latex/figures/}{./}}

\title{Some Tokens Think, Others Remember: \\ Separating Compute and Capacity in LLMs}

\author{Markus Frey\thanks{Lamarr Institute for Machine Learning and Artificial Intelligence, Fraunhofer IAIS, University of Bonn. Correspondence to: \texttt{markus.frey@iais.fraunhofer.de}.} , Behzad Shomali, Joachim Koehler, Mehdi Ali \\
Lamarr Institute, Fraunhofer IAIS, University of Bonn \\
}

\iclrfinalcopy % Uncomment for camera-ready version, but NOT for submission.
\begin{document}
\maketitle
\lhead{} % Clears the "Published as a conference paper at ICLR 2027" text
\renewcommand{\headrulewidth}{0pt} % Optional: removes the horizontal header line
\begin{abstract}
Looped transformers apply a shared block multiple times and have emerged as a parameter-efficient route to scaling compute in language models. However, at fixed FLOPs a looped model has strictly less capacity than a baseline transformer. We argue that \emph{compute}, the number of sequential operations applied to a hidden state, and \emph{capacity}, the parameters available at a single step, are distinct scaling axes that standard transformer layers conflate. We propose a \emph{dual-path block} that exposes both axes as parallel pathways within a single layer: a \emph{deep} sublayer re-applied $K$ times with shared parameters, and a \emph{wide} sublayer with an enlarged feed-forward network applied once, combined by independent per-token gates. At two FLOP budgets, the best dual-path configuration surpasses both iso-FLOP single-axis controls on aggregate language modelling, commonsense, and math metrics, while using 17--33\% \emph{fewer} parameters than the width-scaled baseline. The learned gates are directly interpretable and show systematic per-token allocation: function words and lexical content trend wide, while punctuation, symbols, and arithmetic tokens trend deep. Our results suggest that compute and capacity can be scaled separately, with the model balancing the two token by token.
\end{abstract}
 
% -----------------------------
% INTRODUCTION
% -----------------------------
\section{Introduction}
\label{sec:intro}

Looped (or recursive) transformers re-apply a shared block $K$
times, trading parameter count for sequential compute
\citep{dehghani-etal-2019-ut, geiping-etal-2025-recurrent, saunshi-etal-2025-looped, zhu2025scaling}. The appeal is parameter efficiency:
$L$ layers looped $K$ times reach the effective depth of $KL$
unshared layers at $1/K$ the parameters, and recent work shows
this is enough to recover much of the reasoning performance of
a deeper unshared stack \citep{saunshi-etal-2025-looped, zhu2025scaling}. However, at fixed FLOPs, a looped model has strictly
less capacity than an unshared stack of comparable compute,
and the gap shows up empirically on tasks that depend on
stored knowledge \citep{frey-etal-2026-adaptive, zhu2025scaling}.

Looping and width scaling therefore sit on two qualitatively
different axes of a transformer layer. Looped models use \emph{compute} to increase the number of sequential operations applied to a hidden state, while width-scaled models increase \emph{capacity}, the parameters available at a single step. Standard architectures conflate the two, with every token paying the same cost on both. A looped model puts its whole per-layer feed-forward network (FFN) budget on compute and a width-scaled FFN puts it all on capacity. Neither lets a token that needs more sequential refinement get it without also paying for capacity it does not need, or the other way around.

Recent work relaxes this one axis at a time.
Mixture-of-Experts \citep{shazeer-etal-2017-moe,
fedus-etal-2022-switch} routes tokens to a subset of similar
experts, scaling capacity sub-linearly in compute.
Mixture-of-Depths \citep{raposo-etal-2024-mod} routes tokens to
skip or apply a layer, varying effective depth per position.
Mixture-of-Recursions \citep{bae-etal-2025-mor} varies the
number of loop iterations per token, and adaptive halting
mechanisms in looped models do the same via learned stopping \citep{graves-2016-act, banino2021pondernet,
frey-etal-2026-adaptive}. Each of these routes along a single axis, i.e. which experts, how many loop iterations, or whether to apply a layer at all.

We propose a transformer layer that exposes the two axes
separately within itself. Each \emph{dual-path block} contains
two parallel sublayers: a \emph{deep} sublayer applied $K$
times with shared parameters (a loop, as above), and
a \emph{wide} sublayer with an enlarged feed-forward dimension
applied once. A learned per-token gate combines them. We train the dual-path block across different wide and deep ratios ($\alpha$) and the iso-FLOP controls for each axis separately. We show that: 

\begin{itemize}
\item At both FLOP budgets, the best dual-path configuration
beats both single-axis controls on language-modelling, commonsense, and math evaluations
\emph{while using fewer parameters than the width-scaled
control}.
\item The gate does not collapse to one path. Its allocation depends on the layer and the task. The middle of the stack leans wide and the last layers lean deep. Function words and lexical content (verbs, adjectives) lean wide, while punctuation, symbols, and arithmetic tokens lean deep.
\item The per-token routing affects the output. Forcing arithmetic tokens through the wide path raises GSM8K loss while leaving TriviaQA unchanged, while forcing
verbs, adjectives, and function words through the deep path hurts TriviaQA but not GSM8K.
\end{itemize}

% ---- Two-tone palette (swapped) ----
\definecolor{slate}{HTML}{1F2A44}
\definecolor{slateMid}{HTML}{56607A}
\definecolor{deepC}{HTML}{C7734C}        % warm terracotta — deep / compute
\definecolor{deepCfill}{HTML}{FBE4D6}
\definecolor{deepCfillLight}{HTML}{FDEDE5}
\definecolor{wideC}{HTML}{4C8CB8}        % steel blue — wide / capacity
\definecolor{wideCfill}{HTML}{DDEBF2}
\definecolor{wideCfillLight}{HTML}{EEF5F9}

\pgfdeclarelayer{behind}
\pgfsetlayers{behind,background,main}

% --- Exposed Horizontal Center Position Variables ---
\def\xstd{-5.6}         % Center of Column 1 (Standard Block)
\def\xmid{-2.2}         % Center of Column 2 (Loop-only & Wide-only Blocks)
\def\dualloc{3.2}       % Center of Column 3 (Dual-path Block)
\def\pathx{1.6}

% --- Exposed Loop Offset & Label Position Variables ---
\def\loopwidthb{-0.35}   % Horizontal extension of loop arrow in loop-only block (b)
\def\loopxkbshift{0.1}   % Horizontal shift for \times K text in loop-only block (b)
\def\loopykbshift{0.2}   % Vertical shift for \times K text in loop-only block (b)

\def\loopwidthd{-0.50}   % Horizontal extension of loop arrow in deep path of block (d)
\def\loopxkdshift{0.0}   % Horizontal shift for \times K text in deep path of block (d)
\def\loopykdshift{0.25}   % Vertical shift for \times K text in deep path of block (d)

% Container widths
\def\wstd{2.4}          % Width of Column 1 container
\def\wmid{2.8}          % Width of Column 2 container (Loop/Wide-only)
\def\wdual{6.8}         % Width of Column 3 container (Dual-path)

% Container heights
\def\hstd{7.02}         % Height of Column 1 & Column 3 containers
\def\hmid{3.2}          % Height of Column 2 containers

% --- Exposed Position Variables for Layout Adaptation ---
% Y coordinates for topmost and bottommost text/signals
\def\iny{6.5}
\def\outy{0.1}

% Y coordinates for outer container boundaries
\def\frametop{6.81}
\def\framebottom{-0.21}

% Y coordinates for components in standard block (a)
\def\stdattny{4.4}
\def\stdffny{2.2}

% Y coordinates for components in loop-only block (b)
\def\loopattny{5.65}
\def\loopffny{4.75}
\def\loopouty{3.9}

% Y coordinates for components in wide-only block (c)
\def\wideiny{2.7}
\def\wideattny{1.85}
\def\wideffny{0.95}

% Y coordinates for components in dual-path block (d)
\def\dualbranchy{5.9}
\def\dualhdin{4.8}
\def\dualattny{3.85}
\def\dualffny{2.8}
\def\dualhdout{1.77}
\def\dualinnerframey{5.40}
\def\dualcombiner{1.00}

\begin{figure*}[t]
\centering
\resizebox{0.82\linewidth}{!}{%
\begin{tikzpicture}[
    sblock/.style={
        draw=slate, fill=white, line width=0.4pt,
        minimum width=1.8cm, minimum height=0.6cm,
        rounded corners=2pt,
        font=\footnotesize\bfseries, text=slate
    },
    pathframe/.style={ rounded corners=6pt, line width=0.8pt },
    sumnode/.style={
        draw=slate, fill=white, circle, line width=0.6pt,
        minimum size=0.5cm, inner sep=0pt,
        font=\small, text=slate
    },
    hidden/.style={font=\small, text=slate},
    arr/.style={-{Stealth[length=2mm]}, thick, color=slate},
    arrDeep/.style={-{Stealth[length=2mm]}, thick, color=deepC},
    arrWide/.style={-{Stealth[length=2mm]}, thick, color=wideC},
    looparr/.style={-{Stealth[length=2mm]}, thick, color=deepC},
    title/.style={font=\bfseries\small, text=slate},
    pathtitle/.style={font=\bfseries\small},
    sublabel/.style={font=\itshape\scriptsize, text=slateMid},
    axislabel/.style={font=\scriptsize, text=slateMid},
    smalllabel/.style={font=\scriptsize, text=slateMid},
]

% =========================================================
% COLUMN 1: Standard Transformer Block
% =========================================================
\node[hidden] (xstd_in) at (\xstd, \iny) {$\mathbf{h}^{(\ell-1)}$};
\node[sblock, draw=slate, fill=slate!5] (stdAttn) at (\xstd, \stdattny) {Attn};
\node[sblock, draw=slate, fill=slate!5] (stdFFN) at (\xstd, \stdffny) {FFN ($d_{\text{ffn}}$)};
\node[hidden] (xstd_out) at (\xstd, \outy) {$\mathbf{h}^{(\ell)}$};

\draw[arr] (xstd_in) -- (stdAttn);
\draw[arr] (stdAttn) -- (stdFFN);
\draw[arr] (stdFFN) -- (xstd_out);

\begin{scope}[on background layer]
    \node[pathframe, draw=slate!40, fill=slate!2, minimum width=\wstd cm, minimum height=\hstd cm, anchor=north] (stdframe) at (\xstd, \frametop) {};
\end{scope}
\node[pathtitle, text=slate, anchor=south] at ($(stdframe.north) + (0, 0.05)$) {(a) Standard Block};

% =========================================================
% COLUMN 2 (TOP): Loop-only Block
% =========================================================
\node[hidden] (xloop_in) at (\xmid, \iny) {$\mathbf{h}^{(\ell-1)}$};
\node[sblock, draw=deepC, fill=deepCfillLight] (loopAttn) at (\xmid, \loopattny) {Attn};
\node[sblock, draw=deepC, fill=deepCfillLight] (loopFFN) at (\xmid, \loopffny) {FFN ($d_{\text{deep}}$)};
\node[hidden] (xloop_out) at (\xmid, \loopouty) {$\mathbf{h}^{(\ell)}$};

\draw[arrDeep] (xloop_in) -- (loopAttn);
\draw[arrDeep] (loopAttn) -- (loopFFN);
\draw[arrDeep] (loopFFN) -- (xloop_out);

\draw[looparr, rounded corners=4pt]
  (loopFFN.west) -- ++(\loopwidthb, 0)
  node[midway, above, xshift=\loopxkbshift cm, yshift=\loopykbshift cm, font=\scriptsize, text=deepC!60!black] {$\times K$}
  |- (loopAttn.west);

\begin{scope}[on background layer]
    \node[pathframe, draw=deepC!40, fill=deepCfillLight!20, minimum width=\wmid cm, minimum height=\hmid cm, anchor=north] (loopframe) at (\xmid, \frametop) {};
\end{scope}
\node[pathtitle, text=deepC!80!black, anchor=south] at ($(loopframe.north) + (0, 0.05)$) {(b) PureLoop};

% =========================================================
% COLUMN 2 (BOTTOM): Wide-only Block
% =========================================================
\node[hidden] (xwide_in) at (\xmid, \wideiny) {$\mathbf{h}^{(\ell-1)}$};
\node[sblock, draw=wideC, fill=wideCfillLight] (wideAttn) at (\xmid, \wideattny) {Attn};
\node[sblock, draw=wideC, fill=wideCfillLight, minimum width=2.3cm] (wideFFN) at (\xmid, \wideffny) {FFN ($d_{\text{wide}}$)};
\node[hidden] (xwide_out) at (\xmid, \outy) {$\mathbf{h}^{(\ell)}$};

\draw[arrWide] (xwide_in) -- (wideAttn);
\draw[arrWide] (wideAttn) -- (wideFFN);
\draw[arrWide] (wideFFN) -- (xwide_out);

\begin{scope}[on background layer]
    \node[pathframe, draw=wideC!40, fill=wideCfillLight!20, minimum width=\wmid cm, minimum height=\hmid cm, anchor=south] (wideframe) at (\xmid, \framebottom) {};
\end{scope}
\node[pathtitle, text=wideC!80!black, anchor=south] at ($(wideframe.north) + (0, 0.0)$) {(c) PureWide};

% =========================================================
% COLUMN 3: Dual-path Block  (WIDE on left, DEEP on right)
% =========================================================
\node[hidden] (xdual_in) at (\dualloc, \iny) {$\mathbf{h}^{(\ell-1)}$};
\coordinate (branch) at (\dualloc, \dualbranchy);

% WIDE PATH (left, steel blue)
\node[sblock, draw=wideC, fill=wideCfillLight] (wAttn) at (\dualloc-\pathx, \dualattny) {Attn};
\node[sblock, draw=wideC, fill=wideCfillLight, minimum width=2.3cm] (wFFN) at (\dualloc-\pathx, \dualffny) {FFN ($d_{\text{wide}}$)};
\node[hidden] (hw_in) at (\dualloc-\pathx, \dualhdin) {$\mathbf{h}^{(\ell-1)}$};
\node[hidden] (hw_out) at (\dualloc-\pathx, \dualhdout) {$\mathbf{h}_{\text{wide}}$};

\draw[arrWide] (hw_in) -- (wAttn);
\draw[arrWide] (wAttn) -- (wFFN);
\draw[arrWide] (wFFN) -- (hw_out);

\begin{scope}[on background layer]
    \node[pathframe, draw=wideC, fill=wideCfill, minimum width=3.0cm, minimum height=3.9cm, anchor=north] (wideonlyframe) at (\dualloc-\pathx, \dualinnerframey) {};
\end{scope}
\node[pathtitle, text=wideC!55!black, fill=slate!2, inner sep=2pt, anchor=center] at ($(wideonlyframe.north) + (0, 0.75)$) {Wide};
\node[sublabel, anchor=north west, text=slateMid] at ($(wideonlyframe.north west) + (0.0, -0.0)$) {capacity};

% DEEP PATH (right, terracotta)
\node[sblock, draw=deepC, fill=deepCfillLight] (dAttn) at (\dualloc+\pathx, \dualattny) {Attn};
\node[sblock, draw=deepC, fill=deepCfillLight] (dFFN) at (\dualloc+\pathx, \dualffny) {FFN ($d_{\text{deep}}$)};
\node[hidden] (hd_in) at (\dualloc+\pathx, \dualhdin) {$\mathbf{h}^{(k-1)}$};
\node[hidden] (hd_out) at (\dualloc+\pathx, \dualhdout) {$\mathbf{h}_{\text{deep}}$};

\draw[arrDeep] (hd_in) -- (dAttn);
\draw[arrDeep] (dAttn) -- (dFFN);
\draw[arrDeep] (dFFN) -- (hd_out);

\draw[looparr, rounded corners=6pt]
  (dFFN.east) -- ++(-\loopwidthd, 0)
  node[midway, above, xshift=\loopxkdshift cm, yshift=\loopykdshift cm, font=\scriptsize, text=deepC!60!black] {$\times K$}
  |- (dAttn.east);

\begin{scope}[on background layer]
    \node[pathframe, draw=deepC, fill=deepCfill, minimum width=3.0cm, minimum height=3.9cm, anchor=north] (deepframe) at (\dualloc+\pathx, \dualinnerframey) {};
\end{scope}
\node[pathtitle, text=deepC!55!black, fill=slate!2, inner sep=2pt, anchor=center] at ($(deepframe.north) + (0, 0.75)$) {Deep};
\node[sublabel, anchor=north east, text=slateMid] at ($(deepframe.north east) + (-0.0, -0.0)$) {compute};

% Input routing
\draw[arr] (xdual_in) -- (branch);
\draw[arr, rounded corners=6pt] (branch) -- ($(branch) + (-\pathx, 0)$) -- (hw_in);
\draw[arr, rounded corners=6pt] (branch) -- ($(branch) + (\pathx, 0)$) -- (hd_in);

% Combiner
\node[sumnode] (sum) at (\dualloc, \dualcombiner) {$+$};
\draw[arrWide, rounded corners=8pt] (hw_out.south) |- (sum.west);
\draw[arrDeep, rounded corners=8pt] (hd_out.south) |- (sum.east);

% Output
\node[hidden] (hout) at (\dualloc, \outy)
  {$\mathbf{h}^{(\ell)} = \textcolor{wideC}{g_w}\,\mathbf{h}_{\text{wide}} + \textcolor{deepC}{g_d}\,\mathbf{h}_{\text{deep}}$};
\draw[arr] (sum.south) -- (hout.north);

\begin{pgfonlayer}{behind}
    \node[pathframe, draw=slate!40, fill=slate!2, minimum width=\wdual cm, minimum height=\hstd cm, anchor=north] (dualframe) at (\dualloc, \frametop) {};
\end{pgfonlayer}
\node[pathtitle, text=slate, anchor=south] at ($(dualframe.north) + (0, 0.05)$) {(d) Dual-path Block};

% =========================================================
% BOTTOM PANEL — per-token allocation
% =========================================================
\draw[slateMid!40, dashed, thick] (-6.8, -0.75) -- (6.6, -0.75);

\node[title, anchor=west] at (-6.6, -1.25) {Per-token allocation};
\node[sublabel, anchor=west] at (-6.6, -1.70)
  {learned gates $(g_w, g_d)$ indicate token preference: function words trend wide $\cdot$ arithmetic and symbolic tokens trend deep};

% Axes
\draw[->, slateMid, thick] (-6.1, -3.9) -- (-6.1, -2.15) node[above, font=\scriptsize, text=slateMid] {1};
\draw[slateMid, thick] (-6.1, -3.9) -- (6.1, -3.9);
\node[font=\scriptsize, text=slateMid, rotate=90] at (-6.45, -3.05) {gate value};
\node[font=\scriptsize, text=slateMid] at (-6.25, -3.9) {0};

% Legend (wide first, then deep)
\begin{scope}[shift={(4.9, -2.30)}]
    \node[draw=wideC, fill=wideC, fill opacity=0.6, minimum width=0.30cm, minimum height=0.22cm, inner sep=0pt] (legW) at (0, 0) {};
    \node[smalllabel, anchor=west, text=slate] at (legW.east) {\,wide ($g_w$)};
    \node[draw=deepC, fill=deepC, fill opacity=0.6, minimum width=0.30cm, minimum height=0.22cm, inner sep=0pt] (legD) at (0, -0.35) {};
    \node[smalllabel, anchor=west, text=slate] at (legD.east) {\,deep ($g_d$)};
\end{scope}

% Tokens
\foreach \i/\tok/\gd/\gw in {
    0/{Maria}/0.62/0.40,
    1/{pays}/0.45/0.58,
    2/{15}/0.86/0.22,
    3/{\euro}/0.78/0.30,
    4/{plus}/0.40/0.60,
    5/{a}/0.18/0.82,
    6/{tip}/0.65/0.42,
    7/{of}/0.20/0.80,
    8/{$\tfrac{1}{5}$}/0.88/0.20,
    9/{,}/0.55/0.50,
    10/{the}/0.18/0.80,
    11/{total}/0.58/0.45,
    12/{$=$}/0.82/0.28,
    13/{18}/0.90/0.18,
    14/{\euro}/0.75/0.32
}{
    \pgfmathsetmacro{\xc}{-5.7 + \i*0.72}
    \pgfmathsetmacro{\dx}{0.14}
    \pgfmathsetmacro{\barW}{0.15}
    \pgfmathsetmacro{\ghd}{-3.9 + \gd*1.55}
    \pgfmathsetmacro{\ghw}{-3.9 + \gw*1.55}
    \draw[draw=wideC, fill=wideC, fill opacity=0.6] (\xc-\dx-\barW/2, -3.9) rectangle (\xc-\dx+\barW/2, \ghw);
    \draw[draw=deepC, fill=deepC, fill opacity=0.6] (\xc+\dx-\barW/2, -3.9) rectangle (\xc+\dx+\barW/2, \ghd);
    \node[font=\scriptsize, text=slate, anchor=base] at (\xc, -4.15) {\tok};
}

\end{tikzpicture}%
}
\caption{Block architectures. (a) Standard transformer block. (b) \textsc{PureLoop}: a shared block re-applied $K$ times. (c) \textsc{PureWide}: a single block with enlarged FFN. (d) Dual-path block, which runs (c) and (b) in parallel on the same input and combines them via two per-token sigmoid gates $g_w, g_d$. Bottom panel: schematic of the learned gates across sequence tokens.}
\label{fig:architectures}
\end{figure*}
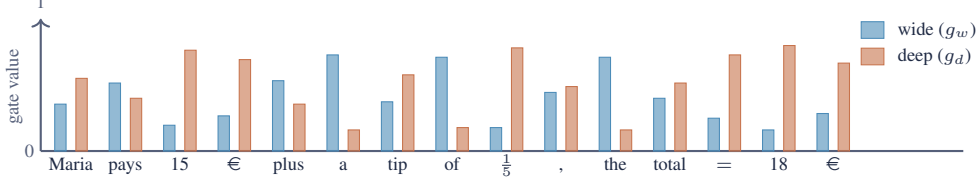

% -----------------------------
% RELATED WORK
% -----------------------------
\section{Related Work}\label{sec:related}
\paragraph{Looped and recursive transformers.}
Re-applying a shared block across depth has been studied as a parameter-efficient route to scaling transformers since the release of Universal Transformers \citep{dehghani-etal-2019-ut}. The idea has been revived recently in language modelling: looped decoders scale test-time compute by re-applying a shared block \citep{geiping-etal-2025-recurrent, saunshi-etal-2025-looped, zhu2025scaling, jeddi2026loopformer, shomali2026loopmtp} which saves parameters while scaling compute. The cost of the parameter saving is reduced capacity. \citet{zhu2025scaling} show that looped models match standard transformers on knowledge manipulation but not on per-parameter memorisation, and \citet{frey-etal-2026-adaptive} report a corresponding empirical pattern downstream: adaptive looping improves mathematical reasoning but leaves commonsense benchmarks largely flat. Both observations show that looping buys compute at the cost of capacity.

\paragraph{Per-token compute allocation.}
A separate line of work allocates compute adaptively at the token level. Mixture-of-Experts \citep{shazeer-etal-2017-moe, fedus-etal-2022-switch} routes each token to a small subset of many parallel experts of the same shape, decoupling parameter count from per-token compute. Mixture-of-Depths \citep{raposo-etal-2024-mod} routes tokens to skip or apply a layer, varying effective depth per position. Mixture-of-Recursions \citep{bae-etal-2025-mor} extends this to looped stacks by varying the number of shared-block applications per token. VersatileFFN \citep{nie2025versatileffn} and MoUE \citep{chen2026mixture} also expose both axes per token. However, both VersatileFFN and MoUE enforce strict parameter reuse (mathematically slicing a shared FFN or reusing a universal expert pool), whereas our dual-path block introduces dedicated capacity parameters via a separate, enlarged wide FFN. Second, both rely on conditional routing to maintain a fixed activation budget. In contrast, our updates are observed at every token and every layer, making our gate's value a direct, un-collapsed read-out of how the trained model chose to allocate its budget between two options that were both processed. 

While MoE and our dual-path block both attach a learned router to a transformer layer, they route over different sets of options. In MoE, the router selects among $N$ feed-forward experts that share the same architecture but hold different learned weights; the choice is over which parameters a token sees. In the dual-path block, the router weighs two sublayers that differ in kind: one is a shared-parameter sublayer applied $K$ times, the other is a wider sublayer applied once. The choice is instead over what type of update a token receives. Furthermore, MoE routers are top-$k$ sparse, whereas our gate is dense. Because both paths are always evaluated in our block, we do not skip compute but re-allocate it. This also means the gate's value is a direct read-out of how the trained model chose to allocate between two options that were both processed. Finally, because the two mechanisms target different axes, they are compositional rather than competing. In theory, a MoE layer could be placed inside the wide path of a dual-path block, with the gate selecting whether to route a token through looped compute or through routed capacity.

\paragraph{Memory-augmented transformers.}
A complementary route to recovering capacity in parameter-shared models is to attach learned memory banks the model can query, as in product-key memory layers \citep{lample2019large} and persistent memory in attention \citep{sukhbaatar2019augmenting}. \citet{frey-etal-2026-adaptive} combine adaptive looping with per-layer and global memory banks and find that memory closes part of the commonsense gap that looping alone cannot bridge. The dual-path block addresses the same capacity bottleneck from the architectural side: rather than adding a separately queried memory module that scales weakly, it adds a parallel wide FFN sublayer inside each layer and lets a per-token gate decide how much of each token's update comes from looped compute vs.\ wider parameters.
 
% =====================================================================
% Methods section — revision 3.
% =====================================================================

\section{Method}
\label{sec:method}

\subsection{Problem statement and notation}
\label{sec:method:problem}

Scaling a transformer layer generally involves either expanding \emph{capacity} (adding parameters, typically via a wider feed-forward network) or extending \emph{compute} (increasing sequential operations on the hidden state, usually via more layers or by recursively re-applying a shared one). We study an architecture that exposes the two axes \emph{separately within each layer} and let the model decide, per token, how to allocate between them.

\subsection{Baseline}
\label{sec:method:backbone}

Given an input tensor $x \in \mathbb{R}^{B \times T \times d}$, where $B$ denotes the batch size, $T$ represents the sequence length, and $d$ is the model dimension, a standard transformer sublayer $\Phi$ is defined as
\begin{align}
\label{eq:sublayer}
u &= x + s \cdot \mathrm{Attn}\!\left(\mathrm{RMSNorm}(x)\right), \nonumber\\
\Phi(x; s) &= u + s \cdot \mathrm{FFN}\!\left(\mathrm{RMSNorm}(u)\right),
\end{align}
where $s \in \mathbb{R}_{>0}$ is a scalar gain on the sublayer
contribution. In a standard transformer $s = 1$; in our dual-path
block, $s$ is learned per recursion step and per path.

\subsection{Dual-path block}
\label{sec:method:dualpath}

Each block exposes the two scaling axes as two parallel sublayers that share the same input $x$, as illustrated in Figure~\ref{fig:architectures}. One sublayer adds sequential compute by re-applying itself $K$ times with shared parameters; the other adds capacity by using a wider FFN and is applied once. A per-token gate combines them. 

\paragraph{Deep path (compute).}
The deep sublayer $\Phi_{\text{deep}}$ uses the FFN hidden dimension $d_{\text{deep}}$ and is applied $K$ times iteratively with shared parameters:
\begin{equation}
\label{eq:deep}
h^{(k)} = \Phi_{\text{deep}}\!\left(h^{(k-1)};\, s^{(k)}_d\right),
\end{equation}
where $h^{(0)} = x$ and $k = 1,\dots,K$. The per-step gains
$s^{(k)}_d = \mathrm{softplus}(\alpha^{(k)})$ are learned, with one
$\alpha^{(k)}$ per step. Initialising $\alpha^{(k)} = -7$ gives
$s^{(k)}_d \approx 9 \times 10^{-4}$, so the recursion begins as a
near-identity and the model learns step-wise deviations from $x$.

Rather than returning $h^{(K)}$ directly, the deep path is a
learned weighted combination of all intermediate states. A small
router (a linear projection from the current state $h^{(k)}$ and a normalised step index $k/(K-1)$) produces a per-step weight $q_k \in [0,1]$. Letting
$\pi_k = \prod_{j<k}(1 - q_j)$, the deep representation is
\begin{equation}
\label{eq:deep-mix}
h_{\text{deep}} \;=\; \sum_{k=1}^{K-1} \pi_k\, q_k\, h^{(k)}
                    \;+\; \pi_K\, h^{(K)}.
\end{equation}
The router thus lets the model down-weight later loop iterations on a per-token basis. 

\paragraph{Wide path (capacity).}
The wide sublayer $\Phi_{\text{wide}}$ has the same attention
configuration as $\Phi_{\text{deep}}$ but an enlarged FFN hidden
dimension $d_{\text{wide}} > d_{\text{deep}}$, and is applied once:
\begin{equation}
\label{eq:wide}
h_{\text{wide}} = \Phi_{\text{wide}}(x;\, s_w), \quad s_w = \mathrm{softplus}(\beta).
\end{equation}
The scalar $\beta$ is initialised to $-7$, matching the deep path.
This path adds parameters through the wider FFN but no sequential
compute beyond a normal layer.

\paragraph{Per-token gating.}
A linear projection $W_g \in \mathbb{R}^{d \times 2}$ with bias
$b_g \in \mathbb{R}^{2}$ maps the layer input to logits
$(\ell_d, \ell_w) = x W_g + b_g$, giving two independent sigmoid
gates $g_d = \sigma(\ell_d)$ and $g_w = \sigma(\ell_w)$, both in
$[0,1]^{B \times T}$. The combined update is
\begin{equation}
\label{eq:two-gates}
y = g_d \odot h_{\text{deep}} \;+\; g_w \odot h_{\text{wide}}.
\end{equation}
We initialise $W_g$ and $b_g$ to zero, so each token receives
$g_d = g_w = 0.5$ at the start of training. The gate is the mechanism by which the model can route tokens that benefit from compute toward the deep path and tokens that benefit from capacity toward the wide path. Because $g_d$ and $g_w$ are independent, their sum can exceed 1. While this is stable at $L=16$ it can cause the residual stream to compound and diverge at greater depths ($L \ge 32$). In the appendix we show ablations using a gate soft cap that normalizes this sum to 1 (see Table \ref{tab:softcap}).

% \rev{We emphasise that this is \emph{not} conditional computation: both paths and all $K$ recursion steps are evaluated for every token, so the gates change each path's contribution to the residual stream but do not skip computation or reduce per-token FLOPs. Dense evaluation is what makes the gate values directly interpretable (Section~\ref{sec:method:readouts}); realising token-dependent FLOP savings would require sparsifying the gate and is left to future work.}

\paragraph{Single-axis baselines.}
Disabling one path reduces a block to a \emph{looped} (\textsc{PureLoop}) layer ($y = h_{\text{deep}}$, $K$ recursions of the standard FFN; Figure~\ref{fig:architectures}b) or a \emph{width-scaled} (\textsc{PureWide}) layer ($y = h_{\text{wide}}$, one pass of the enlarged FFN; Figure~\ref{fig:architectures}c). These two configurations share the backbone, data, and recipe with the dual-path block and put the entire per-layer FFN FLOP budget on one axis. \textsc{PureWide} has the largest parameter count within a budget (no parameter sharing across recursion); \textsc{PureLoop} has the smallest (one FFN re-applied $K$ times). 

% \paragraph{A dense, sparsity-free gate.}
% Per-token routing also exists in mixture-of-experts (MoE)
% \citep{shazeer-etal-2017-moe, fedus-etal-2022-switch},
% mixture-of-depths (MoD) \citep{raposo-etal-2024-mod}, and
% mixture-of-recursions (MoR) \citep{bae-etal-2025-mor}, but those
% gates are sparse: each token visits one or a small top-$k$ subset
% of branches. Sparsity confounds the measurement we want. If a
% token is routed to expert 3, the absence of a contribution from
% experts 1, 2, 4 is a routing decision, not evidence about what
% that token ``needed'' from those branches; the unselected experts
% are simply not evaluated. Sparse routing also typically requires
% an auxiliary load-balancing loss \citep{fedus-etal-2022-switch,
% zoph-etal-2022-stmoe}. Our gate is dense: both paths are evaluated for
% every token, and $g_d, g_w \in [0,1]$ are independent sigmoids
% trained from the LM loss alone. The gate's value at a position
% therefore records what the trained model chose to do with both
% options available.

\subsection{Routing read-outs}
\label{sec:method:readouts}
 
Because both paths are evaluated for every token
(Section~\ref{sec:method:dualpath}), the forward pass exposes the
raw gates ($(g_d, g_w)$) \emph{and} the per-path update vectors
$\Delta_d = h_{\text{deep}} - x$ and
$\Delta_w = h_{\text{wide}} - x$ at every layer and every token.

The raw gate value alone ignores update size. We therefore report the fraction of the residual update that came from the deep path,
\begin{equation}
\label{eq:deep-share}
\rho_d \;=\; \frac{g_d\,\|\Delta_d\|}{g_d\,\|\Delta_d\| + g_w\,\|\Delta_w\|}
\;\in\; [0,1],
\end{equation}
computed per token per layer. $\rho_d = 1$ means the entire
update at that position came from the deep path; $\rho_d = 0$
means it came entirely from the wide path; $\rho_d = 0.5$ is
balanced. We refer to $\rho_d$ as the \emph{deep share} throughout.

% \rev{Note that because the two sigmoid gates are independent, they control
% both the relative weighting of the paths and the overall scale of
% the residual update. $\rho_d$ should therefore be read as a ratio of
% gate-weighted update \emph{norms}, i.e.\ a relative-allocation
% statistic, rather than an exact decomposition of the final residual
% contribution (the two updates are vectors and can partially cancel
% or reinforce; Figure~\ref{fig:path_alignment}b quantifies their
% alignment).}
 
\paragraph{Path alignment.}
We also record the cosine similarity between the two path
deltas,
\begin{equation}
\label{eq:path-cos}
\cos(\Delta_d, \Delta_w) \;=\;
\frac{\Delta_d \cdot \Delta_w}{\|\Delta_d\|\,\|\Delta_w\|}.
\end{equation}
A value near $+1$ means the two paths push the residual in the
same direction, i.e. deep and wide path do the same, while a value near $0$ means they push in orthogonal directions.
 
% \paragraph{Task-aligned differences.}
% For Figure~\ref{fig:sequence_grid}(c) we align $\rho_d$ from two
% sources (GSM8K and TriviaQA) to a shared anchor token
% (\texttt{Answer}) and report the per-layer, per-offset difference
% $\rho_d^{\text{GSM8K}} - \rho_d^{\text{TriviaQA}}$ averaged over
% matched chunks. Positive values indicate offsets where GSM8K
% routes more deeply than TriviaQA at that layer.
 
% All four quantities are computed at inference time on a held-out
% slice of Paloma (WikiText-103, TriviaQA, GSM8K) and require no
% modification to the trained model. They are dense by
% construction: every token and every layer contributes to every
% quantity, so aggregations by layer, by part-of-speech, or by task
% position do not need to control for routing sparsity.

% -----------------------------
% EXPERIMENTS
% -----------------------------
\definecolor{midC}{HTML}{8E6F96}
\colorlet{oursrow}{midC!15!white}

\section{Experiments}
\label{sec:experiments}

The backbone is a GPT2-style \citep{radford2019language} decoder-only transformer with
$L = 16$ layers, hidden dimension $d = 768$, and $12$ attention heads. Attention
uses rotary positional embeddings \citep{su-etal-2024-rope}
applied to queries and keys, with RMSNorm on queries and keys
prior to attention \citep{dehghani-etal-2023-vit22b}. Feed-forward sublayers are SwiGLU
\citep{shazeer-2020-glu}. These parameters are held fixed across every configuration in the paper,
including the single-axis controls; only the FFN widths and the
per-layer recursion depth vary. 
 
\subsection{Setup}
\label{sec:exp:setup}

\paragraph{Models and configurations.}
We train models at two iso-FLOP budgets, specified by the per-layer FFN FLOPs per token ($F_M = 80$M and
$F_M = 160$M, corresponding to $\sim$1.28G and $\sim$2.56G total FLOPs per token for a 16-layer model). The detailed procedure for solving for matched FFN widths is given in Appendix~\ref{sec:app:flopmatch}.  Within each budget we sweep the dual-path FFN allocation $\alpha \in \{25, 50, 75\}$ (the share of FFN FLOPs spent on the deep path) and the recursion depth $K \in \{2, 3, 4\}$.  At fixed budget, larger $K$ and
larger $\alpha$ reduce parameter count, since both shift compute toward the shared-parameter deep path.  Unique parameter counts across all configurations span from roughly 240M up to 1.4B.  All models share pre-RMSNorm \citep{zhang-sennrich-2019-rmsnorm}, RoPE \citep{su-etal-2024-rope}, SwiGLU \citep{shazeer-2020-glu} and QK-norm \citep{henry-etal-2020-qknorm}.  Exact $L$, $d$, $h_q$, $h_{kv}$, $d_{\text{ffn}}$,
$d_{\text{ffn}}^{\text{wide}}$ per configuration are listed in
Appendix~\ref{sec:app:configs} (Table~\ref{tab:configs-f80m}).

\paragraph{Data and training.}
All models are trained on a deduplicated subset of Nemotron-CC
\citep{su-etal-2024-nemotron-cc} for 38B tokens with the GPT-2
tokenizer.  Sequence length is $4096$.  We use the
\texttt{modalities} training framework \citep{lubbering2026modalities} using the AdamW optimizer (with peak learning rate $5\times 10^{-4}$, a linear warmup of $184$ steps, and a cosine decay schedule down to $5\times 10^{-5}$). Wall-clock training times range from 12.8 to 21.4 hours per model across 64 GPUs. Full optimizer hyperparameters are listed in Appendix~\ref{sec:app:training}.

\paragraph{Baselines.}
We compare against two iso-FLOP single-axis controls, trained
with the same backbone, data, and recipe:

\begin{itemize}
\item \textsc{PureWide} ($L$ layers, single wider FFN of width
$d_{\text{ffn}}^{\text{wide}}$, no recursion): our block with the
deep path disabled.  Spends the entire per-layer FFN FLOP budget
on capacity.
\item \textsc{PureLoop} ($L$ layers, standard FFN width,
recursion depth $K$, no wide path): our block with the wide path
disabled.  Spends the entire per-layer FFN FLOP budget on
sequential compute.
\end{itemize}

\paragraph{Evaluation.}
We report bits-per-byte (BPB) on Paloma C4 and
WikiText-103 for language modelling, mean accuracy and BPB on six
commonsense tasks (ARC-c, ARC-e, HellaSwag, PIQA, SIQA,
WinoGrande), GSM8k accuracy and the OLMo3 base-easy math BPB
average over seven sub-tasks, and LAMBADA / QASPER for reading
and QA.  BPB rather than accuracy gives finer-grained signal
for these pre-training scales. The aggregate BPB used in
Section~\ref{sec:exp:main} is computed hierarchically: we first
form a commonsense BPB (the unweighted mean over the six
commonsense task BPBs) and a math BPB (the unweighted mean over
the seven OLMo3 base-easy math sub-task BPBs), then take the
unweighted mean of four category values: C4, WikiText-103,
commonsense, and math.  Full per-task numbers are in Appendix~\ref{sec:app:eval}
(Tables~\ref{tab:appendix-f80m} and \ref{tab:appendix-f160m}).
 
\subsection{Main results}

\begin{table}[t]
\centering
\caption{\textbf{Main results.} Iso-FLOP comparison at two budgets ($\text{Mean} \pm \text{Std}$ across three seeds). \textsc{Dual} configurations use \texttt{loop=4}; rows vary the FFN FLOP allocation between the deep and wide paths ($a25/a50/a75$ = 25/50/75\% on the deep path). The full sweep over loop depths is in Appendix~\ref{sec:app:configs}. Params is shown relative to \textsc{PureWide} within each budget. All rows within a budget are matched in FLOPs.}
\label{tab:main}
\footnotesize
\setlength{\tabcolsep}{3pt}
\renewcommand{\arraystretch}{1.0}
\resizebox{\textwidth}{!}{%
\begin{tabular}{l c c cc cc cc}
\toprule
 & & & \multicolumn{2}{c}{Language modelling} & \multicolumn{2}{c}{Commonsense} & \multicolumn{2}{c}{Math} \\
\cmidrule(lr){4-5}\cmidrule(lr){6-7}\cmidrule(lr){8-9}
Configuration & Params & FLOPs & C4 $\downarrow$ & Wiki $\downarrow$ & Acc. $\uparrow$ & BPB $\downarrow$ & Acc. $\uparrow$ & BPB $\downarrow$ \\
\midrule
\textsc{PureWide} & 719M (1.00$\times$) & 80M & 1.037\std{0.002} & 0.946\std{0.003} & 0.5030\std{0.0204} & 0.9965\std{0.0083} & 0.0732\std{0.0164} & 0.5305\std{0.0039} \\
\multicolumn{9}{l}{\textsc{PureLoop}} \\
\quad Loop=2 & 398M (0.55$\times$) & 80M & 1.051\std{0.002} & 0.961\std{0.002} & 0.4962\std{0.0093} & 1.0034\std{0.0023} & 0.0762\std{0.0160} & 0.5382\std{0.0059} \\
\quad Loop=3 & 294M (0.41$\times$) & 80M & 1.065\std{0.001} & 0.976\std{0.002} & 0.5038\std{0.0056} & 1.0256\std{0.0045} & 0.0580\std{0.0130} & 0.5500\std{0.0081} \\
\quad Loop=4 & 238M (0.33$\times$) & 80M & 1.078\std{0.001} & 0.992\std{0.002} & 0.4991\std{0.0071} & 1.0334\std{0.0024} & 0.0566\std{0.0122} & 0.5632\std{0.0060} \\
\rowcolor{oursrow}
\multicolumn{9}{l}{\textsc{DualPath} \emph{(ours)}} \\
\rowcolor{oursrow}
\quad $\alpha=25$ & 606M (0.84$\times$) & 80M & 1.037\std{0.002} & 0.942\std{0.003} & 0.5061\std{0.0134} & 0.9914\std{0.0025} & 0.0659\std{0.0047} & 0.5229\std{0.0024} \\
\rowcolor{oursrow}
\quad $\alpha=50$ & 483M (0.67$\times$) & 80M & \textbf{1.036}\std{0.001} & \textbf{0.939}\std{0.002} & \textbf{0.5192}\std{0.0081} & \textbf{0.9816}\std{0.0004} & \textbf{0.0982}\std{0.0073} & \textbf{0.5200}\std{0.0047} \\
\rowcolor{oursrow}
\quad $\alpha=75$ & 360M (0.50$\times$) & 80M & 1.045\std{0.001} & 0.948\std{0.002} & 0.5043\std{0.0232} & 0.9955\std{0.0101} & 0.0905\std{0.0118} & 0.5289\std{0.0057} \\
\midrule
\textsc{PureWide} & 1361M (1.00$\times$) & 160M & 1.018\std{0.001} & 0.921\std{0.002} & 0.5376\std{0.0032} & 0.9682\std{0.0089} & 0.1195\std{0.0050} & 0.5066\std{0.0010} \\
\multicolumn{9}{l}{\textsc{PureLoop}} \\
\quad Loop=2 & 719M (0.53$\times$) & 160M & 1.027\std{0.000} & 0.929\std{0.002} & 0.5294\std{0.0033} & 0.9821\std{0.0045} & 0.1201\std{0.0147} & 0.5095\std{0.0010} \\
\quad Loop=3 & 502M (0.37$\times$) & 160M & 1.037\std{0.000} & 0.940\std{0.002} & 0.5215\std{0.0027} & 0.9846\std{0.0050} & 0.1126\std{0.0260} & 0.5141\std{0.0010} \\
\quad Loop=4 & 398M (0.29$\times$) & 160M & 1.045\std{0.002} & 0.946\std{0.002} & 0.5134\std{0.0014} & 0.9973\std{0.0113} & 0.0957\std{0.0138} & 0.5190\std{0.0052} \\
\rowcolor{oursrow}
\multicolumn{9}{l}{\textsc{DualPath} \emph{(ours)}} \\
\rowcolor{oursrow}
\quad $\alpha=25$ & 1125M (0.83$\times$) & 160M & \textbf{1.014}\std{0.001} & \textbf{0.913}\std{0.002} & \textbf{0.5393}\std{0.0040} & \textbf{0.9607}\std{0.0033} & 0.1182\std{0.0064} & 0.5069\std{0.0024} \\
\rowcolor{oursrow}
\quad $\alpha=50$ & 880M (0.65$\times$) & 160M & 1.017\std{0.001} & 0.917\std{0.003} & 0.5380\std{0.0048} & 0.9705\std{0.0128} & 0.1250\std{0.0083} & 0.5019\std{0.0076} \\
\rowcolor{oursrow}
\quad $\alpha=75$ & 644M (0.47$\times$) & 160M & 1.021\std{0.001} & 0.921\std{0.002} & 0.5339\std{0.0010} & 0.9655\std{0.0041} & \textbf{0.1442}\std{0.0089} & \textbf{0.5015}\std{0.0039} \\
\bottomrule
\end{tabular}%
}
\end{table}

\label{sec:exp:main}
\paragraph{A dual-path configuration performs best at both budgets.}
We focus on $K{=}4$ throughout this section as it is the deepest recursion in our sweep and therefore yields the largest parameter savings at fixed FLOPs. The full sweep over $K \in \{2,3,4\}$ is reported in Appendix~\ref{sec:app:eval}. At both FLOP budgets, a dual-path configuration beats both single-axis controls on aggregate BPB. At $F_M = 80$M, $\alpha = 50$ ($483$M params, $0.67\times$ \textsc{PureWide}) achieves the best aggregate BPB ($0.8692$ vs.\ $0.8775$ for \textsc{PureWide} and $0.8884$ for the best \textsc{PureLoop}), and is also best on C4, Wiki, commonsense BPB, math BPB, and commonsense accuracy. At $F_M = 160$M, the optimum shifts toward capacity: $\alpha = 25$ ($1125$M params, $0.83\times$ \textsc{PureWide}) is best on aggregate BPB ($0.8487$ vs.\ $0.8535$ for \textsc{PureWide} and $0.8619$ for the best \textsc{PureLoop}), and wins C4, Wiki, commonsense BPB, and commonsense accuracy. This shift toward capacity at the higher budget is consistent with the broader picture that looped (parameter-shared) compute is parameter-bottlenecked.

\paragraph{Allocation between deep and wide path.}
We now investigate how the model performance changes when varying $\alpha$ within the $K{=}4$ models. Capacity-heavy configurations ($\alpha = 25$) are strongest on language modelling and commonsense: at $F_M = 160$M, $\alpha{=}25$ gives the best C4 and WikiText-103 BPB and the best commonsense accuracy and BPB. Compute-leaning configurations are strongest on math: at $F_M = 160$M, GSM8k accuracy and math BPB peak at $\alpha{=}75$ ($0.1442$ and $0.5015$), and at $F_M = 80$M both $\alpha{=}50$ and $\alpha{=}75$ clearly exceed the capacity-heavy $\alpha{=}25$ on GSM8k ($0.0982$ and $0.0905$ vs.\ $0.0659$). This shows that shifting FFN FLOPs toward the wide path adds parameters and helps knowledge-heavy tasks, while shifting them toward the deep path adds sequential compute and helps reasoning-heavy tasks.

\begin{figure}[t]
    \centering
    \includegraphics[width=1.0\linewidth]{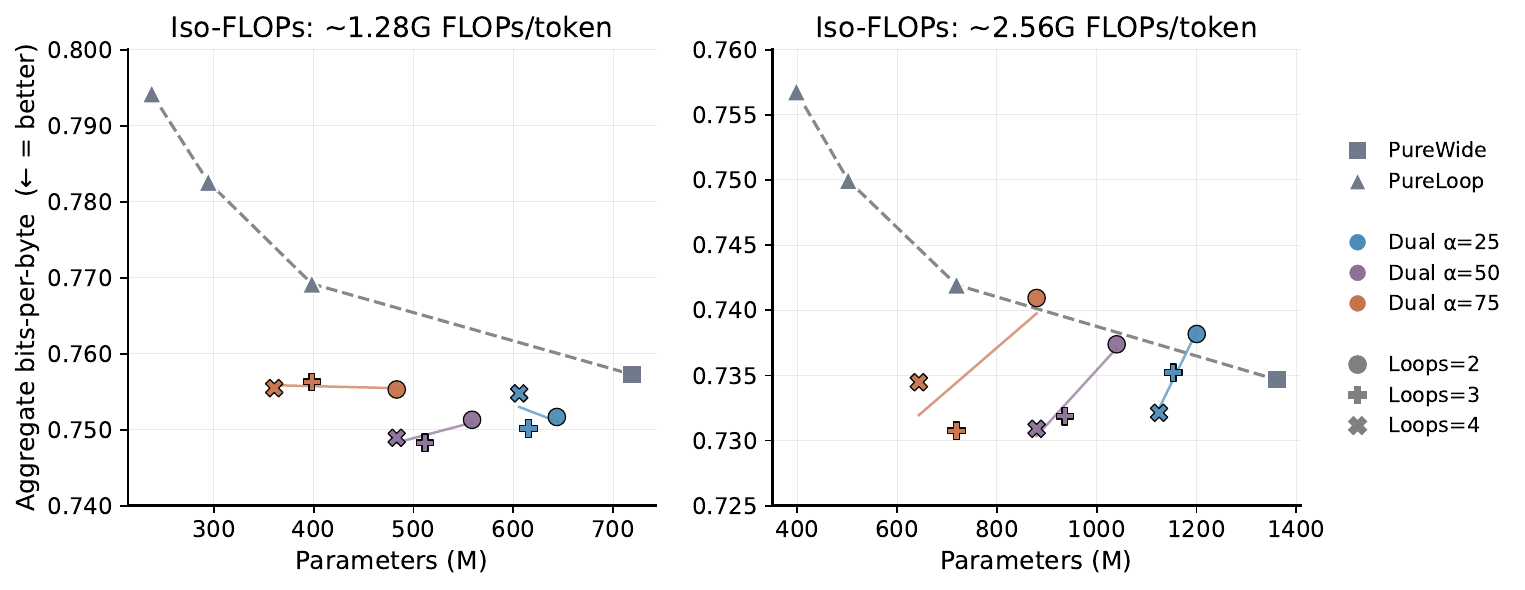}
    \caption{\textbf{Parameter scaling sweep of dual-path configurations} compared against \textsc{PureWide} and \textsc{PureLoop} controls at matched FLOP budgets. Number of Loops for \textsc{PureLoop} (triangle) is K=4, 3, 2 (from left to right). Note that \textsc{PureWide} is equivalent to \textsc{PureLoop} with K=1 but without additional routing overhead.}
    \label{fig:headline_scatter}
\end{figure}

\paragraph{Position in the parameter--quality plane.}
We next plot aggregated bits-per-byte against parameter count for all configurations ($K \in \{2,3,4\}$ and $\alpha \in \{25,50,75\}$) in Figure~\ref{fig:headline_scatter}. For optimal parameter efficiency at a fixed FLOP budget, a model should land in the lower left corner. Within each $\alpha$, connecting $K \in \{2, 3, 4\}$ traces a short per-$\alpha$ segment moving from right to left (as higher $K$ reduces parameter count at fixed FLOPs). At the lower budget ($F_M = 80$M, $\sim$1.28G FLOPs), the segments have different slopes, with some increasing performance along the line while others remain flat. However, at the higher budget ($F_M = 160$M, $\sim$2.56G FLOPs), increasing the loop count generally improves performance across all $\alpha$ values, driving the models further into the optimal lower-left region.

\paragraph{Scaling depth.}
Table~\ref{tab:depth-sweep} keeps the per-layer budget at $F_M = 80$M and varies the number of layers $L \in \{8, 16, 32\}$, using the gate cap $c=1$ for all dual-path runs (without it the $L=32$ run diverges). At $L=8$ and $L=16$, the dual-path block beats \textsc{PureWide} on every metric with $0.67$--$0.70\times$ the parameters. At $L=32$ it still leads on math (GSM8K $0.1748$ vs.\ $0.1260$, math BPB $0.4848$ vs.\ $0.4925$), but \textsc{PureWide} is now slightly ahead on C4, WikiText-103, and commonsense. This suggests that a deep enough stack already supplies most of the sequential compute that language modelling and commonsense need.

\subsection{Where does the model spend its budget?}
\label{sec:exp:analysis}

Having established that the dual-path block improves overall performance, we now examine the learned gates to understand how the model allocates its budget. Because the architecture evaluates both paths densely, we can directly read out the routing decisions and residual updates at inference time to uncover the model's underlying preferences for sequential compute (the deep path) or parameter capacity (the wide path). We find that the routing preference varies a) across the layer's position in the stack, b) the identity of the token (e.g. noun vs. number), and c) the task the token is part of (math vs. question answering). Unless otherwise stated, all analyses below use the
balanced $\alpha = 50$, $K = 4$ model evaluated on three Paloma
sources (WikiText-103, TriviaQA, GSM8K).

\begin{figure}[t]
\centering
\includegraphics[width=\linewidth]{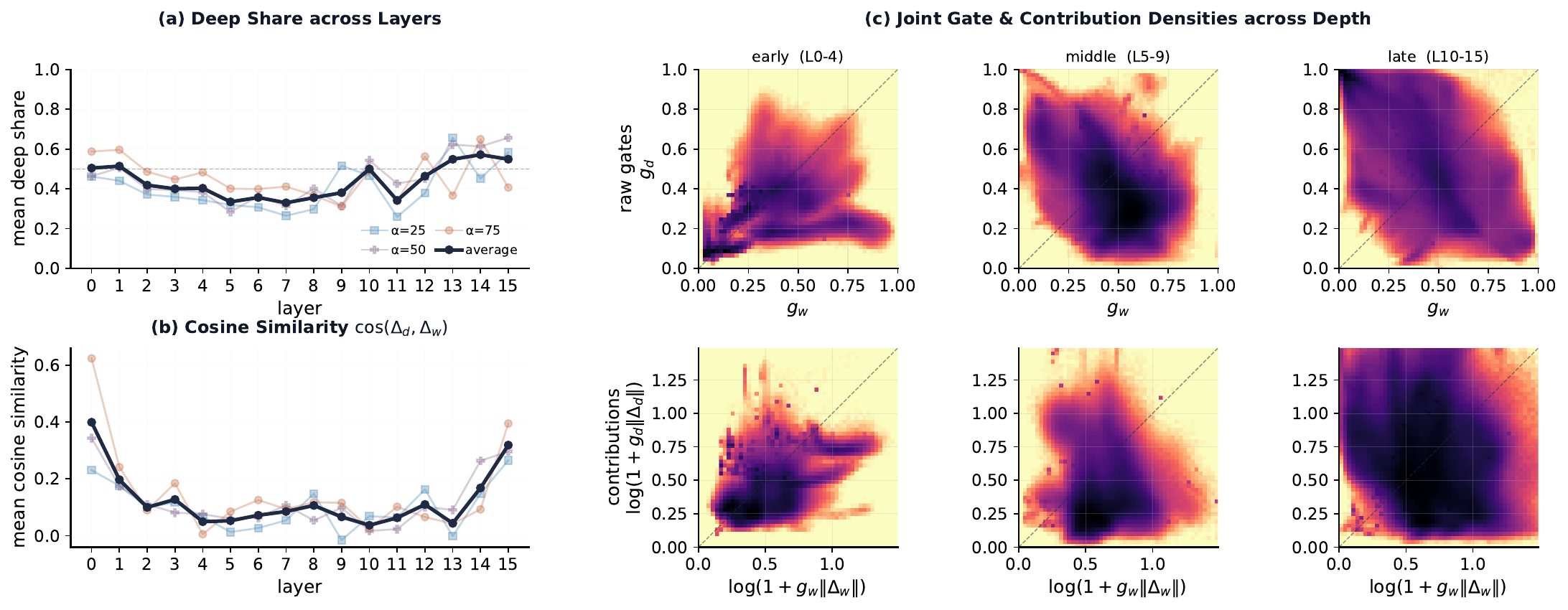}
\caption{\textbf{Routing share, representation alignment, and joint gate densities across depth.} 
(a)~Mean deep share per layer for $\alpha \in \{25, 50, 75\}$ and average. 
(b)~Mean cosine similarity between deep and wide updates ($\Delta_d, \Delta_w$). 
(c)~Joint 2D density across depth bands (early L0--4, middle L5--9, late L10--15): raw gates $(g_w, g_d)$ (top) and update contributions weighted by norm (bottom).}
\label{fig:path_alignment_gate_density}
\label{fig:path_alignment}
\label{fig:gate_density}
\end{figure}

\paragraph{Depth in the stack.}
Figure~\ref{fig:path_alignment}(a) plots the mean deep share per layer for $\alpha \in \{25, 50, 75\}$, along with their average. Two regimes are visible. The middle of the network (L2--L9) is wide-dominated, with deep share between $0.28$ and $0.40$, while the last two layers (L14--L15) flip to deep-dominated, with the average rising to $0.5$--$0.6$. The pattern is consistent across alpha values, so the shape reflects the layer's role in the stack rather than the specific ratio between compute and capacity. 

Panel (b) shows the mean cosine similarity between the update vectors from the deep path ($\Delta_d$) and the wide path ($\Delta_w$) at each layer. A value near $+1$ means the paths produce nearly identical updates (and one is redundant), a value near $0$ means they push in orthogonal directions (the paths contribute non-overlapping information). We observe mostly low values across the middle of the network, indicating that the deep and wide paths are processing the same input in genuinely different ways rather than producing scaled copies of one another.

\paragraph{Task.}
The gate responds to the task a token is part of. Figure~\ref{fig:routing_pref_page_8} shows two example sequences side by side: in the GSM8K answer, numbers and arithmetic operators (\texttt{15}, \texttt{*}, \texttt{=}, \texttt{3}, \texttt{18}) are deep-leaning, and the deep preference strengthens through the arithmetic blocks. In the TriviaQA example, the answer token \texttt{Oxy} is the most wide-leaning position in the sequence, consistent with factual recall drawing on the capacity path. 

Figure~\ref{fig:sequence_grid} resolves the same examples layer by layer and shows the depth pattern from earlier: later layers prefer the deep path, mid layers prefer the wide path. For a population estimate we align one thousand samples from each task to the \texttt{Answer} token and plot the per-layer difference $\rho_d^{\text{GSM8K}} - \rho_d^{\text{TriviaQA}}$ (Figure~\ref{fig:sequence_grid}c). Alignment to a shared anchor is needed because the \texttt{Answer} token sits at different absolute positions in sequences of varying length. The post-\texttt{Answer} positions in GSM8K are markedly more deep-leaning than the corresponding TriviaQA positions in the late layers. At the same depth and the same relative position, a reasoning task routes more deeply than a knowledge task.

\paragraph{Token identity.}
To investigate in more detail beyond task identity we tag every token with its Universal POS tag \citep{petrov2012universal, nivre2016ud} using spaCy \citep{honnibal2020spacy} (\texttt{en\_core\_web\_sm}) on the decoded text, with a regex override for arithmetic tokens (digits, operators, \texttt{=}, \texttt{<<}, \texttt{>>}, \texttt{\#\#\#\#}). Each model token inherits the tag of the character span it overlaps most. Figure~\ref{fig:pos_main} plots, for every (POS, layer) pair, the mean deep share over all tokens with that tag, restricted to tags with at least 10 occurrences. The ordering is stable across all three datasets and visible in the boxplot of panel~(e): SPACE, PUNCT (e.g., \texttt{,}, \texttt{.}), and SYM (e.g., \texttt{=}, \texttt{-}, \texttt{<<}) receive the highest deep share; ADV (e.g., \texttt{also}, \texttt{as}), PART (e.g., \texttt{to}, \texttt{'s}), PRON (e.g., \texttt{he}, \texttt{it}), ADJ (e.g., \texttt{many}, \texttt{first}), and VERB (e.g., \texttt{made}, \texttt{used}) receive the lowest; NUM (e.g., \texttt{2}, \texttt{5}) and NOUN (e.g., \texttt{Question}, \texttt{Answer}) sit in the middle. Overall, the POS pattern follows an interpretable split: symbolic and structural tokens (SPACE, PUNCT, SYM) route to compute, while lexical content tokens (VERB, ADJ, ADV, PRON, PART), route to capacity. When overriding the gates of arithmetic tokens to bypass the deep path, we observe a degradation in GSM8K loss compared to a size-matched random override, while forcing function words through the deep path selectively degrades WikiText-103 and TriviaQA but not GSM8K (Table~\ref{tab:token-class-overrides}).

\paragraph{Per-token decisions become more polarised with depth.}
The analyses so far show \emph{what} the gate prefers but not \emph{how strongly} it commits to those preferences. We therefore plot the joint distribution of the
two gates. Figure~\ref{fig:gate_density} plots the 2D density of $(g_w, g_d)$ pooled across all tokens, grouped into three layer bands (early L0--4, middle L5--9, late L10--15). The dashed diagonal marks equal routing ($g_d = g_w$): mass on the diagonal corresponds to balanced mixtures, while mass off the diagonal
corresponds to tokens that commit primarily to one path. The per-layer breakdown is shown in Appendix Figure~\ref{fig:gate_density_layers}. We observe that the density moves off the diagonal as depth increases. Early layers (L0--4) concentrate in a narrow band along the positive diagonal near the origin, indicating small and roughly balanced gates for most tokens. Late layers (L10--15)
push density onto the boundaries with a cluster near $g_d \to 1$ (tokens routed almost entirely through the deep path) and a cluster near $g_w \to 1$ (tokens routed almost entirely through the wide path). Row~B plots the same density after re-weighting each axis by the actual update magnitude ($\log(1 + g_{w}\|\Delta_{w}\|)$ vs.\ $\log(1 + g_{d}\|\Delta_{d}\|)$), which accounts for the fact that a high gate value contributes little if its path's update is small. We observe that the gate is not just choosing different mixtures for different tokens but it also commits more strongly to one path or the other as the residual stream moves through the network.

% =====================================================================
% Conclusion
% =====================================================================
\section{Conclusion}
\label{sec:conclusion}

Standard transformer layers conflate two distinct scaling axes:
compute (sequential operations on a hidden state) and capacity
(parameters available at a single step). The dual-path block
separates them, using a recursive deep sublayer and a parallel
wider sublayer combined by a dense learned gate. At two iso-FLOP budgets, the best dual-path configuration improves on both single-axis controls on aggregate language modelling, commonsense, and math metrics while using fewer parameters than the width-scaled baseline. The ratio between the wide and deep path $\alpha$ provides a predictable trade-off, with capacity-heavy settings favoring knowledge tasks and compute-heavy settings favoring reasoning.

Our results show that this parallel formulation can outperform iso-FLOP single-axis baselines on aggregate language modelling, commonsense, and math metrics at two FLOP budgets. Because both paths are evaluated at every token, the routing decisions are a direct read-out of how the trained model chose to spend its budget, not a sampling artifact. This learned allocation is interpretable. It varies systematically with layer depth, with wide being preferred in the middle of the stack, and deep dominated in the last two layers. Moreover, function words and lexical content trend wide while punctuation, symbols, and arithmetic tokens trend deep.

The dual-path block opens several directions we find promising. First, the per-$\alpha$ trajectories in Figure~\ref{fig:headline_scatter} steepen from the lower to the higher FLOP budget. Whether this trend continues, plateaus, or reverses at billion-parameter budgets and across different depth/width ratios is the most direct extension.

Second, as noted above, the dual-path gate and MoE routing target orthogonal axes and are architecturally compositional: loops increase FLOPs while keeping parameters the same, while MoEs increase the parameters while keeping FLOPs fixed (with fixed top-k). In our model a MoE layer could occupy the wide path, with the outer gate deciding between looped compute and routed capacity.

% Lastly, the per-token, per-layer gate indicate which positions the model treats as compute-bound versus capacity-bound. Beyond what our analysis show this could be used to identify tokens where reasoning fails. 
These directions suggest that separating compute and capacity within a layer can be an extension that combines with other scaling ideas, while providing an interpretable signal as a free byproduct.

\section{Limitations}
\label{sec:discussion:limitations}

Our experiments cover two per-layer FLOP budgets and models up to 1.36B parameters trained on 38B tokens. The dual-path advantage holds at both budgets and across three backbone depths (Table~\ref{tab:depth-sweep}), and testing whether it persists at larger scales is an open question. The dual-path block does not provide conditional computation: both paths and all $K$ recursion steps run for every token, so the gates re-allocate contribution but do not reduce per-token FLOPs (Section~\ref{sec:method:dualpath}). Sparsifying the gate to obtain per-token savings is left for future work.

The model does not benefit from running more loops at inference than it was trained with. This is expected for fixed training depth and prior work obtains extrapolation by randomising the recursion depth during training \citep{geiping-etal-2025-recurrent}, which is compatible with our block. Lastly, with independent gates, uncapped training diverges at $L \geq 32$. Capping $g_d + g_w$ at 1 removes the divergence and improves results at $L = 16$ (Table~\ref{tab:softcap}).

\subsubsection*{AI Use Statement}
\label{sec:ai_use}
In this manuscript, we used Claude for writing support, i.e. copy-editing and phrasing and for ideation during the intial stages of the project. We also used it for visualization of results and for evaluating weaknesses in the manuscript. We have not used it to generate datasets or for designing the experiments. All AI-assisted text and code were reviewed and verified for correctness. We take full responsibility for the final content of this work. 

\bibliographystyle{iclr2027_conference}
\bibliography{custom}
% -----------------------------
% APPENDIX STUBS
% -----------------------------
\clearpage
\appendix
\section*{Appendix}
\addcontentsline{toc}{section}{Appendix}

\section{Training Details}
\label{sec:app:training}

All models are trained using the \texttt{modalities} framework. We use the AdamW optimizer with $\beta_1 = 0.9$, $\beta_2 = 0.95$, $\epsilon = 10^{-8}$, and weight decay of $0.3$ (excluding embedding and RMSNorm layers). The learning rate is warmed up linearly from $5\times 10^{-6}$ to a peak of $5\times 10^{-4}$ over $184$ steps, and then decayed to $5\times 10^{-5}$ using a cosine annealing schedule. Wall-clock training times range from 12.8 to 21.4 hours per model (average of 16.0 hours) on the cluster hardware. Total pretraining tokens are 38B.

\section{FLOP-matching protocol}
\label{sec:app:flopmatch}
The FLOP budget $F_M$ is the per-token, per-layer FFN compute. For each configuration we solve for the largest legal $d_{\text{ffn}}$ (and, for dual-path, $d_{\text{ffn}}^{\text{wide}}$) whose induced $h_{\text{eff}}$ keeps the per-layer FFN cost at or below $F_M$, also accounting for the router on the dual-path block. Within each budget the controls and the dual configurations agree on total per-token FFN FLOPs to within $0.5\%$; the same backbone ($L = 16$, $d = 768$, $h_q = h_{kv} = 12$) is held fixed across all configurations, so attention compute is identical.

This appendix gives the exact accounting used to match FLOPs
across configurations. All FLOPs are reported per token and per layer.

\paragraph{Per-sublayer FLOPs.}
Let $d$ be the model dimension, $n_{\text{rep}} = h_q / h_{kv}$
the repeat factor (1 in all our runs), and $d_{\text{ffn}}$
the \emph{configured} FFN hidden width. The SwiGLU \emph{effective}
hidden width used by the model is
\begin{equation}
\label{eq:heff}
h_{\text{eff}}(d_{\text{ffn}}) \;=\; 64 \cdot
\Bigl\lceil \tfrac{1}{64} \lfloor 2 d_{\text{ffn}} / 3 \rfloor \Bigr\rceil,
\end{equation}
i.e.\ the LLaMA-style $2/3$ scaling rounded up to a multiple of 64.
The per-token FLOPs of one sublayer are
\begin{align}
\mathrm{FLOP}_{\text{attn}} &= 4 d^2 + 4 d^2 / n_{\text{rep}}, \\
\mathrm{FLOP}_{\text{ffn}}(d_{\text{ffn}}) &= 6 d \cdot h_{\text{eff}}(d_{\text{ffn}}).
\end{align}
The dual-path two-gate router contributes
$\mathrm{FLOP}_{\text{gate}} = 4 d$ FLOPs per token (linear
$d \to 2$).

\newcommand{\tokenbox}[2]{%
  \colorbox[rgb]{#1}{\strut\footnotesize\texttt{#2}}%
}
\setlength{\fboxsep}{1.0pt}

\newcommand{\routingcolorbar}{%
\begin{tikzpicture}
  % The gradient bar
  \shade[left color={rgb,1:red,0.445;green,0.692;blue,0.827},
         middle color={rgb,1:red,1.0;green,1.0;blue,1.0},
         right color={rgb,1:red,0.992;green,0.859;blue,0.780}]
    (0,0) rectangle (6,0.25);
  % Border
  \draw[black,line width=0.3pt] (0,0) rectangle (6,0.25);
  % Labels
  \node[anchor=east, font=\footnotesize] at (-0.1,0.125) {wide-leaning};
  \node[anchor=west, font=\footnotesize] at (6.1,0.125) {deep-leaning};
\end{tikzpicture}}
% ========================================== Page 8 ==========================================
\begin{figure}[t]
\centering
\begin{minipage}{0.95\textwidth}
  \subsection*{GSM8K Mathematical Reasoning Example}
  {\setlength{\baselineskip}{11pt}%
   \tokenbox{0.978,0.923,0.891}{Question}\tokenbox{0.991,0.863,0.788}{:} \tokenbox{0.810,0.893,0.938}{Ashley}\tokenbox{0.752,0.864,0.922}{'s} \tokenbox{0.606,0.790,0.880}{pizza} \tokenbox{0.645,0.809,0.891}{delivery} \tokenbox{0.810,0.893,0.938}{costs} \tokenbox{0.831,0.904,0.943}{\$}\tokenbox{0.849,0.912,0.947}{15}\tokenbox{0.942,0.956,0.964}{.} \tokenbox{0.781,0.878,0.930}{What} \tokenbox{0.843,0.909,0.945}{is} \tokenbox{0.791,0.883,0.933}{the} \tokenbox{0.771,0.873,0.927}{total} \tokenbox{0.878,0.926,0.952}{amount} \tokenbox{0.942,0.956,0.964}{that} \tokenbox{0.781,0.878,0.930}{Ashley} \tokenbox{0.606,0.790,0.880}{should} \tokenbox{0.800,0.888,0.936}{give} \tokenbox{0.636,0.805,0.889}{the} \tokenbox{0.626,0.800,0.886}{delivery} \tokenbox{0.810,0.893,0.938}{man} \tokenbox{0.907,0.940,0.957}{if} \tokenbox{0.616,0.795,0.883}{she} \tokenbox{0.761,0.869,0.925}{wants} \tokenbox{0.684,0.829,0.902}{to} \tokenbox{0.771,0.873,0.927}{give} \tokenbox{0.626,0.800,0.886}{a} \tokenbox{0.655,0.814,0.894}{tip} \tokenbox{0.606,0.790,0.880}{that} \tokenbox{0.645,0.809,0.891}{is} \tokenbox{0.674,0.824,0.900}{equal} \tokenbox{0.684,0.829,0.902}{to} \tokenbox{0.820,0.898,0.941}{1}\tokenbox{0.781,0.878,0.930}{/}\tokenbox{0.810,0.893,0.938}{5} \tokenbox{0.855,0.915,0.948}{of} \tokenbox{0.742,0.859,0.919}{the} \tokenbox{0.866,0.920,0.950}{amount} \tokenbox{0.597,0.785,0.878}{she} \tokenbox{0.837,0.906,0.944}{ordered}\tokenbox{0.976,0.932,0.906}{?}\newline
   \tokenbox{0.820,0.898,0.941}{Answer}\tokenbox{0.985,0.893,0.839}{:} \tokenbox{0.936,0.953,0.963}{The} \tokenbox{0.925,0.948,0.961}{tip} \tokenbox{0.972,0.954,0.943}{that} \tokenbox{0.901,0.937,0.956}{Ashley} \tokenbox{0.925,0.948,0.961}{wants} \tokenbox{0.831,0.904,0.943}{to} \tokenbox{0.925,0.948,0.961}{give} \tokenbox{0.866,0.920,0.950}{amounts} \tokenbox{0.966,0.967,0.968}{to} \tokenbox{0.936,0.953,0.963}{\$}\tokenbox{0.973,0.949,0.935}{15} \tokenbox{0.969,0.966,0.965}{x} \tokenbox{0.976,0.932,0.906}{1}\tokenbox{0.970,0.962,0.958}{/}\tokenbox{0.977,0.928,0.899}{5} \tokenbox{0.966,0.967,0.968}{=} \tokenbox{0.896,0.934,0.955}{\$}\tokenbox{0.866,0.920,0.950}{\textless{}\textless{}}\tokenbox{0.986,0.889,0.832}{15}\tokenbox{0.976,0.936,0.913}{*}\tokenbox{0.985,0.893,0.839}{1}\tokenbox{0.976,0.932,0.906}{/}\tokenbox{0.992,0.859,0.780}{5}\tokenbox{0.978,0.923,0.891}{=}\tokenbox{0.992,0.859,0.780}{3}\tokenbox{0.984,0.898,0.847}{\textgreater{}\textgreater{}}\tokenbox{0.984,0.898,0.847}{3}\tokenbox{0.989,0.872,0.803}{.}\newline
   \tokenbox{0.665,0.819,0.897}{H}\tokenbox{0.966,0.967,0.968}{ence}\tokenbox{0.984,0.898,0.847}{,} \tokenbox{0.931,0.951,0.962}{she} \tokenbox{0.866,0.920,0.950}{will} \tokenbox{0.936,0.953,0.963}{give} \tokenbox{0.884,0.928,0.953}{a} \tokenbox{0.849,0.912,0.947}{total} \tokenbox{0.985,0.893,0.839}{of} \tokenbox{0.982,0.906,0.862}{\$}\tokenbox{0.981,0.910,0.869}{15} \tokenbox{0.973,0.949,0.935}{+} \tokenbox{0.981,0.910,0.869}{\$}\tokenbox{0.985,0.893,0.839}{3} \tokenbox{0.975,0.941,0.921}{=} \tokenbox{0.942,0.956,0.964}{\$}\tokenbox{0.942,0.956,0.964}{\textless{}\textless{}}\tokenbox{0.978,0.923,0.891}{15}\tokenbox{0.979,0.919,0.884}{+}\tokenbox{0.989,0.872,0.803}{3}\tokenbox{0.977,0.928,0.899}{=}\tokenbox{0.991,0.863,0.788}{18}\tokenbox{0.988,0.876,0.810}{\textgreater{}\textgreater{}}\tokenbox{0.985,0.893,0.839}{18} \tokenbox{0.960,0.964,0.967}{to} \tokenbox{0.925,0.948,0.961}{the} \tokenbox{0.855,0.915,0.948}{delivery} \tokenbox{0.976,0.932,0.906}{man}\tokenbox{0.991,0.851,0.770}{.}\newline
   \tokenbox{0.919,0.945,0.959}{\#\#\#\#} \tokenbox{0.966,0.967,0.968}{18}
  }
  \subsection*{TriviaQA Factual Knowledge Example}
  {\setlength{\baselineskip}{11pt}%
   \tokenbox{0.978,0.923,0.891}{Question}\tokenbox{0.991,0.863,0.788}{:} \tokenbox{0.866,0.920,0.950}{What} \tokenbox{0.890,0.931,0.954}{is} \tokenbox{0.855,0.915,0.948}{the} \tokenbox{0.665,0.819,0.897}{second} \tokenbox{0.606,0.790,0.880}{most} \tokenbox{0.606,0.790,0.880}{common} \tokenbox{0.665,0.819,0.897}{gas} \tokenbox{0.843,0.909,0.945}{in} \tokenbox{0.800,0.888,0.936}{the} \tokenbox{0.925,0.948,0.961}{atmosphere}\tokenbox{0.974,0.945,0.928}{?}\newline
   \tokenbox{0.761,0.869,0.925}{Answer}\tokenbox{0.936,0.953,0.963}{:} \tokenbox{0.445,0.692,0.827}{Oxy}\tokenbox{0.907,0.940,0.957}{gen}
  }
\end{minipage}
\protect\routingcolorbar\\
\caption{Token-level deep share for GSM8K and TriviaQA. Blue denotes wide-leaning (prefers the capacity path) while red denotes deep-leaning, meaning it prefers the compute (looped) path. }
\label{fig:routing_pref_page_8}
\end{figure}

\begin{figure*}[t]
    \centering
    \includegraphics[width=1\linewidth]{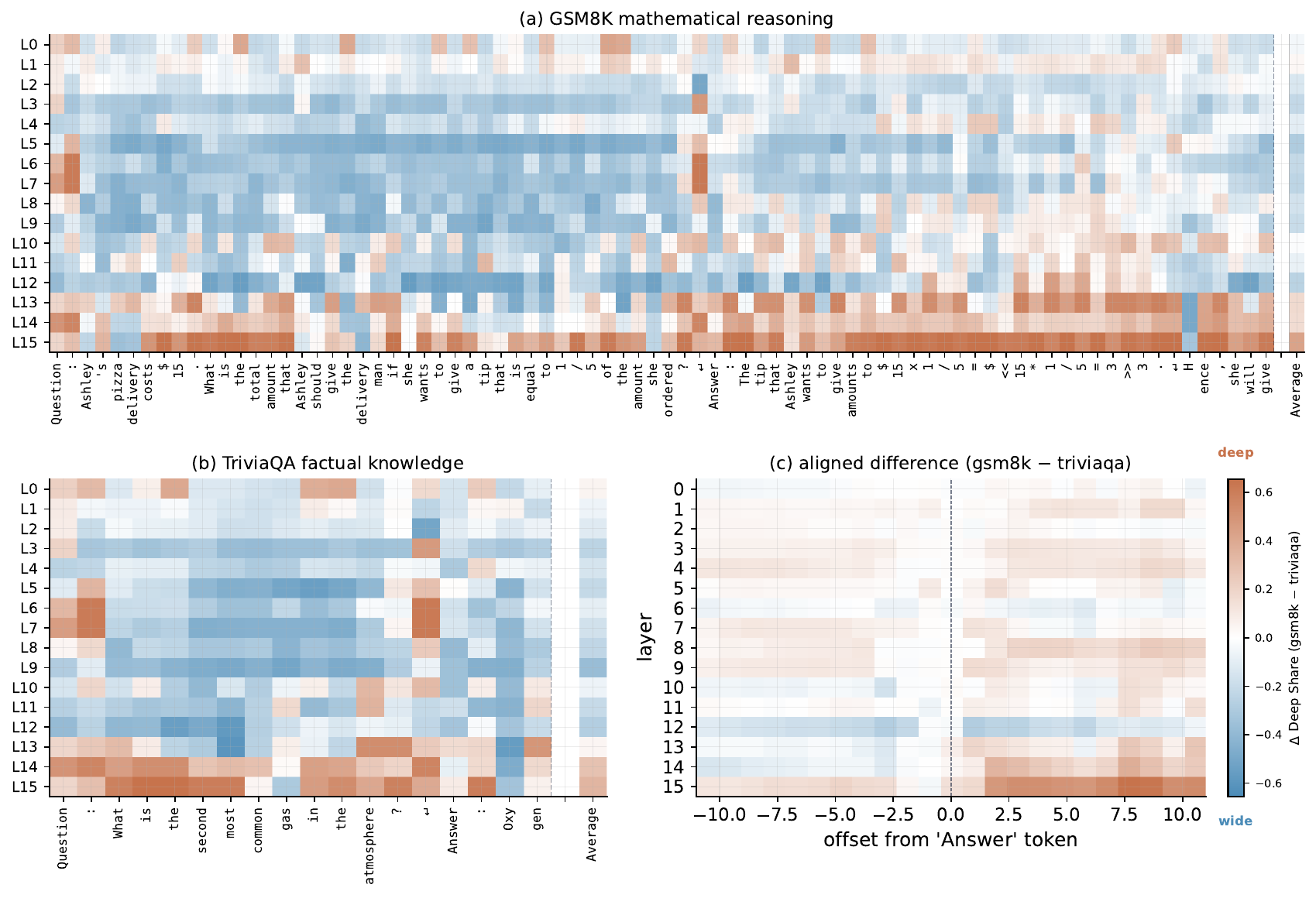}
    \caption{\textbf{Step-by-step token-level routing grid and task
    alignment.} Panel (a) shows the layer-by-token heatmap of the
    deep share for a sequence from GSM8K
    (mathematical reasoning). Panel (b) shows the heatmap
    for a sequence from TriviaQA (factual knowledge). Panel (c) shows the aligned difference in preference ($\text{gsm8k} - \text{triviaqa}$) around the anchor token
    ``Answer'', averaged over one thousand sequences per dataset.}
    \label{fig:sequence_grid}
\end{figure*}

\begin{figure*}[t]
    \centering
    \includegraphics[width=0.95\linewidth]{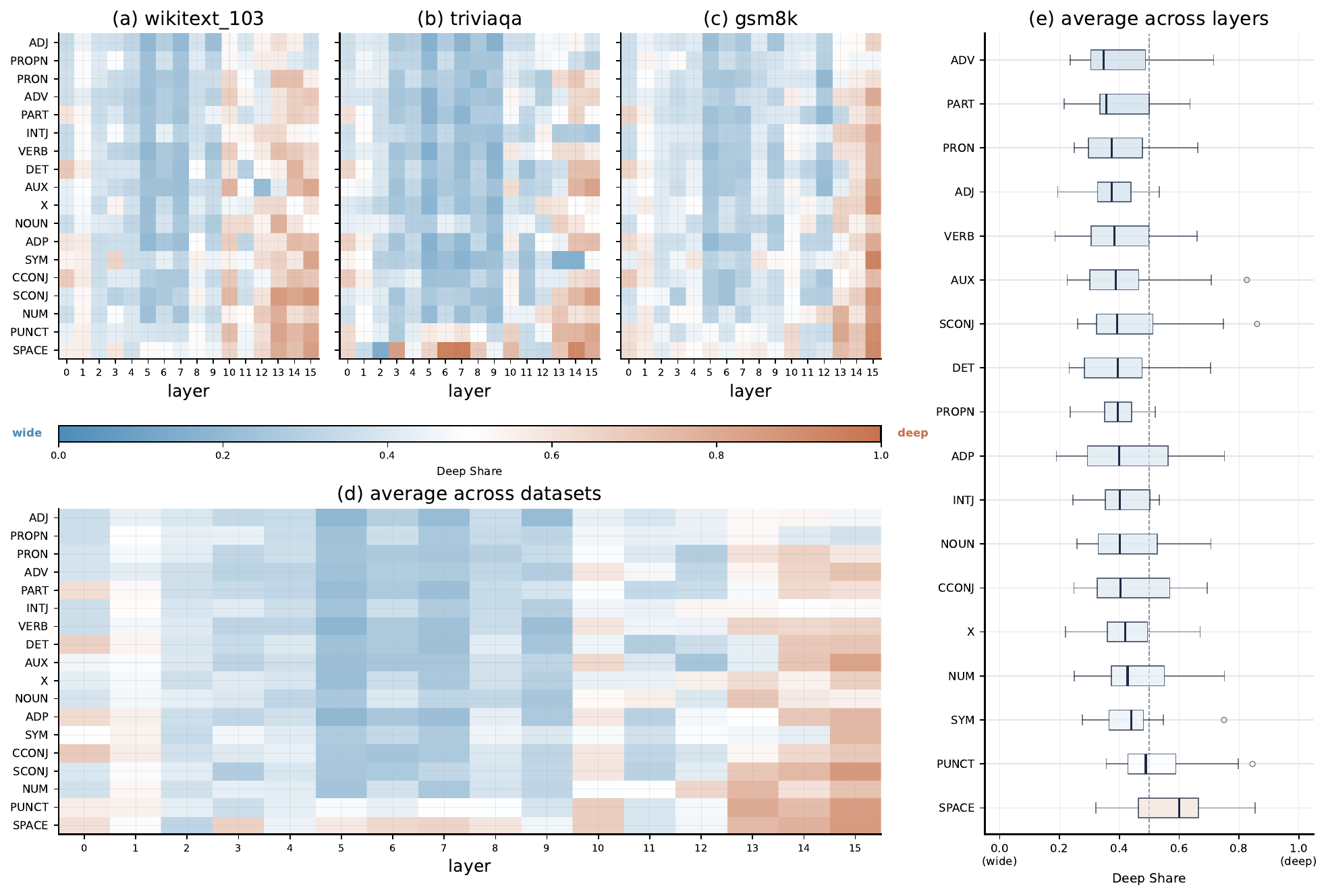}
    \caption{\textbf{Parts-of-speech (POS) routing characteristics and commitment.} Panels (a, b, c) show heatmaps of the mean deep share per universal POS tag across layers for \texttt{wikitext\_103}, \texttt{triviaqa}, and \texttt{gsm8k}, respectively, with tags sorted by the overall mean deep share. Panel (d) plots the average heatmap across the three datasets. Panel (e) shows boxplots of the deep share distribution per POS tag across all layers and datasets, sorted by their median preference (most deep-leaning at the bottom).}
    \label{fig:pos_main}
\end{figure*}

\paragraph{Per-layer FLOP budgets.}
For the three layer types, the per-layer FFN-side FLOP count
$F_M$ (which we hold fixed at $80$M or $160$M) is
\begin{align}
F_M^{\textsc{PureWide}}
  &= \mathrm{FLOP}_{\text{attn}}
   + \mathrm{FLOP}_{\text{ffn}}(d_{\text{ffn}}^{\text{wide}}), \\
F_M^{\textsc{PureLoop}}
  &= K \bigl(\mathrm{FLOP}_{\text{attn}}
   + \mathrm{FLOP}_{\text{ffn}}(d_{\text{ffn}})\bigr), \\
F_M^{\textsc{Dual}}
  &= K \bigl(\mathrm{FLOP}_{\text{attn}}
   + \mathrm{FLOP}_{\text{ffn}}(d_{\text{ffn}})\bigr) \nonumber \\
  &\quad{}+ \mathrm{FLOP}_{\text{attn}}
   + \mathrm{FLOP}_{\text{ffn}}(d_{\text{ffn}}^{\text{wide}}) \nonumber \\
  &\quad{}+ \mathrm{FLOP}_{\text{gate}}.
\end{align}
The dual-path layer therefore pays an extra attention pass and a
small router term relative to a single-axis layer at the same
FFN FLOPs. Note that we hold all FLOPs fixed at 80M or 160M per layer and per token by adjusting the size of the FFN.

\paragraph{Solving for FFN widths.}
Given $F_M$, $K$, and (for dual) a deep-FLOP fraction
$\alpha \in (0,1)$, we set
$\mathrm{FLOP}_{\text{ffn}}(d_{\text{ffn}})
  = (\alpha (F_M - \mathrm{FLOP}_{\text{gate}}) / K)
  - \mathrm{FLOP}_{\text{attn}}$
and analogously for the wide width with $(1-\alpha)$. We then
invert Eq.~\ref{eq:heff} for the \emph{largest} $d_{\text{ffn}}$
(rounded down to a multiple of 64) whose $h_{\text{eff}}$ keeps
the equation at or below the target -- the ``floor'' rounding
mode. The exception is \textsc{PureWide}, where we round up
(``ceil''); this keeps the largest single-axis capacity baseline
honest by spending the entire budget and has strictly more FLOPs than the dual-path baselines. The residual mismatch is $<2\%$ of
$F_M$ in every configuration. 

\newcommand{\ph}{\textcolor{gray}{\textsc{tbd}}}

\begin{table}[t]
\centering
\caption{\textbf{Gate soft cap.} The two-gate residual gain $g_d+g_w$ lies in $(0,2)$ and is free
to sit above $1$, so it compounds with depth. Capping the \emph{sum} at $c=1$ (dividing both gates
by $\max(1,(g_d+g_w)/c)$, which preserves the routing ratio $g_d\!:\!g_w$ exactly and leaves tokens
below the cap untouched) removes the divergence seen at $L\!\ge\!32$ and does not cost accuracy at
$L=16$. All rows are $\alpha=50$, \texttt{loop=4}, FLOP-matched to \textsc{PureWide}. Single runs.}
\label{tab:softcap}
\small
\setlength{\tabcolsep}{4pt}
\begin{tabular}{l c c cc cc cc}
\toprule
 & & & \multicolumn{2}{c}{Language modelling} & \multicolumn{2}{c}{Commonsense} & \multicolumn{2}{c}{Math} \\
\cmidrule(lr){4-5}\cmidrule(lr){6-7}\cmidrule(lr){8-9}
Configuration & Params & FLOPs & C4 $\downarrow$ & Wiki $\downarrow$ & Acc. $\uparrow$ & BPB $\downarrow$ & Acc. $\uparrow$ & BPB $\downarrow$ \\
\midrule
\textsc{PureWide} & 719M & 80M & 1.037 & 0.946 & 0.5203 & 0.9920 & 0.0791 & 0.5261 \\
\textsc{DualPath} $\alpha=50$ & 483M & 80M & \textbf{1.036} & 0.939 & \textbf{0.5256} & 0.9820 & 0.0898 & 0.5198 \\
\rowcolor{oursrow}
\quad + gate cap $c=1$ & 483M & 80M & \textbf{1.036} & \textbf{0.933} & 0.5233 & \textbf{0.9740} & \textbf{0.1348} & \textbf{0.5116} \\
\midrule
\textsc{PureWide} & 1361M & 160M & 1.018 & 0.922 & 0.5384 & 0.9636 & 0.1250 & 0.5077 \\
\textsc{DualPath} $\alpha=50$ & 880M & 160M & 1.017 & 0.920 & 0.5326 & 0.9704 & 0.1328 & 0.4959 \\
\rowcolor{oursrow}
\quad + gate cap $c=1$ & 880M & 160M & \textbf{1.010} & \textbf{0.905} & \textbf{0.5430} & \textbf{0.9529} & \textbf{0.1377} & \textbf{0.4894} \\
\bottomrule
\end{tabular}
\end{table}

\begin{table}[t]
\centering
\caption{\textbf{Depth sweep at the F80M budget.} $F_M$ is a \emph{per-layer} budget, so the block
is identical at every depth and only $L$ changes; total FLOPs/token and parameters scale linearly
with $L$. \textsc{DualPath} uses the gate cap $c=1$, \texttt{loop=4}. Without the cap the $L=32$ run diverges to \texttt{NaN}, so to make this sweep comparable we re-ran L=8 \& 16 using the cap $c=1$.}
\label{tab:depth-sweep}
\small
\setlength{\tabcolsep}{4pt}
\begin{tabular}{ll c c cc cc cc}
\toprule
 & & & & \multicolumn{2}{c}{Language modelling} & \multicolumn{2}{c}{Commonsense} & \multicolumn{2}{c}{Math} \\
\cmidrule(lr){5-6}\cmidrule(lr){7-8}\cmidrule(lr){9-10}
$L$ & Configuration & Params & GFLOPs & C4 $\downarrow$ & Wiki $\downarrow$ & Acc. $\uparrow$ & BPB $\downarrow$ & Acc. $\uparrow$ & BPB $\downarrow$ \\
\midrule
 & \textsc{PureWide} & 398M & 0.64 & 1.078 & 0.997 & 0.4893 & 1.0469 & 0.0322 & 0.5724 \\
\rowcolor{oursrow}
\multirow{-2}{*}{8}
 & \textsc{DualPath} $\alpha=50$ & 280M & 0.63 & \textbf{1.072} & \textbf{0.983} & \textbf{0.4992} & \textbf{1.0252} & \textbf{0.0508} & \textbf{0.5611} \\
\midrule
 & \textsc{PureWide} & 719M & 1.28 & 1.037 & 0.946 & 0.5203 & 0.9920 & 0.0791 & 0.5261 \\
\rowcolor{oursrow}
\multirow{-2}{*}{16}
 & \textsc{DualPath} $\alpha=50$ & 483M & 1.26 & \textbf{1.036} & \textbf{0.933} & \textbf{0.5233} & \textbf{0.9740} & \textbf{0.1348} & \textbf{0.5116} \\
\midrule
 & \textsc{PureWide} & 1361M & 2.57 & \textbf{1.008} & \textbf{0.903} & \textbf{0.5438} & \textbf{0.9488} & 0.1260 & 0.4925 \\
\rowcolor{oursrow}
\multirow{-2}{*}{32}
 & \textsc{DualPath} $\alpha=50$ & 889M & 2.52 & 1.014 & 0.905 & 0.5412 & 0.9517 & \textbf{0.1748} & \textbf{0.4848} \\
\bottomrule
\end{tabular}
\end{table}

\section{Model Configurations}
\label{sec:app:configs}

All configurations share the same backbone: $L = 16$ layers,
$d = 768$, $h_q = h_{kv} = 12$ (GQA repeat factor 1), sequence
length 4096, vocabulary size 50{,}304, weight-tied input/output
embeddings, and the SwiGLU effective-hidden rule of
Appendix~\ref{sec:app:flopmatch} (multiple of 64). Table
\ref{tab:configs-f80m} lists, for
every model in this paper, the configured deep FFN width
$d_{\text{ffn}}$, wide FFN width
$d_{\text{ffn}}^{\text{wide}}$, recursion depth $K$, total
parameter count, and wall-clock training time. 

\section{Inference-time ablations}
\label{sec:app:ablations}

We probe the trained dual-path model ($F_M = 80$M, $\alpha = 50$,
$K = 4$) at inference time, overriding either the loop count or
the gate.  Table~\ref{tab:ablations} reports per-token
cross-entropy on three Paloma sources.

\paragraph{The loop budget matters, and the schedule does not
extrapolate.}
Forcing fewer loop iterations at inference monotonically degrades
loss (e.g.\ GSM8K loss rises from $2.148$ at the trained $K{=}4$
to $3.592$ at $K{=}1$).  Returns diminish past the trained value, i.e. the loop dynamics learned at training do not extrapolate past the training budget. 

\paragraph{Both paths are needed.}
Disabling either gate ($g_d{=}1, g_w{=}0$ or $g_d{=}0, g_w{=}1$)
costs $3.6$--$5.6$ nats across the three sources -- larger than
the gap between an early-training and a fully-trained model.  A
uniform $0.5{:}0.5$ split still loses $\sim 2.6$ nats.  Opening
both gates fully ($g_d{=}1, g_w{=}1$) is the worst override: the
model has trained with bounded gate magnitudes and the residual
update is far out of its training distribution.

\paragraph{The per-token decision carries information.}
The shuffled-gates row keeps the marginal distribution of
$(g_d, g_w)$ but randomises which token gets which assignment
within each sequence.  Loss rises by $0.51$--$0.94$ nats, indicating that the
gate's per-token decisions, not just its average behaviour are
load-bearing.

\paragraph{Routing choices are functionally necessary for specific token types.}
The global overrides above show that the gate matters overall. To test if individual tokens rely on their preferred paths, we run gate overrides across all checkpoints. We push arithmetic tokens (\textsc{num} and \textsc{sym}) through the wide path instead of the deep one, and function words (\textsc{adj}, \textsc{adv}, \textsc{part}, \textsc{pron}, \textsc{verb}) through the deep path instead of the wide one. Because any forced routing hurts a trained model, we compare each intervention against a control group where an identical number of random tokens are overridden. Table~\ref{tab:token-class-overrides} reveals that forcing arithmetic tokens wide causes a consistent loss spike on GSM8K ($+0.37$ nats over control, $p<0.001$) and WikiText-103, but leaves TriviaQA unaffected. Conversely, forcing function words deep hurts language modeling on WikiText-103 and TriviaQA ($p<0.05$) while sparing math.

\section{Evaluation Details}
\label{sec:app:eval}

We evaluate language modelling on Paloma C4 and WikiText-103
(bits-per-byte).  Commonsense is the mean over ARC-c, ARC-e,
HellaSwag, PIQA, SIQA, and WinoGrande.  Math is GSM8k accuracy
and the OLMo3 base-easy math BPB average over Algebra, Counting,
Geometry, Intermediate Algebra, Number Theory, Pre-algebra, and
Pre-calculus. The aggregate BPB reported in Section~\ref{sec:exp:main} is the unweighted mean of four category values (C4, WikiText-103, the six-task commonsense BPB mean, and the seven-sub-task math BPB mean), so each category contributes equally. 
% Note that for Figure \ref{fig:headline_scatter} we instead average across BPB values from \emph{all} benchmarks as listed in Table~\ref{tab:appendix-f80m} ($F_M = 80$M) and Table~\ref{tab:appendix-f160m} ($F_M = 160$M).
% Evaluations are run using the OLMES evaluation framework \citep{gu2025olmes}.

\begin{figure*}[t]
    \centering
    \includegraphics[width=1\linewidth]{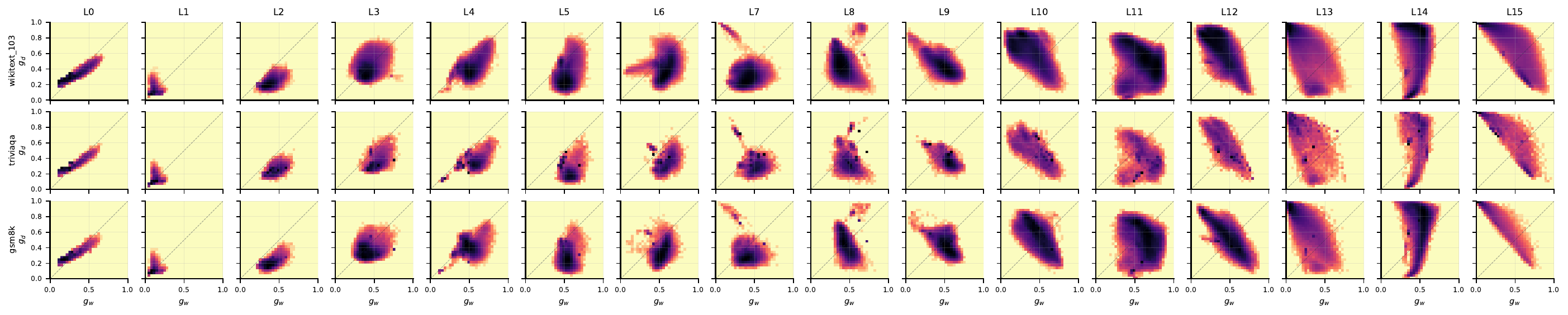}
    \caption{\textbf{Layer-wise joint density of raw gates} ($g_w, g_d$) across all layers ($L0$ to $L15$) and evaluation sources. The rows correspond to the three Paloma evaluation datasets (\texttt{wikitext\_103}, \texttt{triviaqa}, and \texttt{gsm8k}), while columns correspond to layers index 0 to 15. The diagonal dashed line in each subplot represents equal routing preference ($g_d = g_w$). Lighter regions represent higher token concentration. The model has $L=16$ layers, $K=4$ loops, $d_{\text{model}} = 768$, deep FFN hidden width $= 4864$, wide FFN hidden width $= 24448$, dual $\alpha=50$.}
    \label{fig:gate_density_layers}
\end{figure*}

\begin{table*}[t]
\centering
\small
\setlength{\tabcolsep}{4pt}
\renewcommand{\arraystretch}{1.05}
\begin{tabular}{lccc}
\toprule
\textbf{Intervention} & \textbf{GSM8K} & \textbf{TriviaQA} & \textbf{Wiki-103} \\
\midrule
\multicolumn{4}{l}{\emph{Loss increase over learned gates ($\Delta$ CE, mean $\pm$ sd)}} \\
\quad Arith.\ forced wide            & 2.30 \ssub{$\pm$0.35} & 0.14 \ssub{$\pm$0.02} & 0.67 \ssub{$\pm$0.13} \\
\quad Random ctrl (wide)             & 1.93 \ssub{$\pm$0.31} & 0.14 \ssub{$\pm$0.02} & 0.41 \ssub{$\pm$0.07} \\
\quad Func.\ words forced deep       & 0.75 \ssub{$\pm$0.22} & 0.71 \ssub{$\pm$0.19} & 1.08 \ssub{$\pm$0.24} \\
\quad Random ctrl (deep)             & 0.73 \ssub{$\pm$0.14} & 0.65 \ssub{$\pm$0.13} & 0.94 \ssub{$\pm$0.11} \\
\midrule
\multicolumn{4}{l}{\emph{Class-specific excess over random control}} \\
\quad Arith.\ forced wide            & $+0.37^{***}$ & $+0.00$ & $+0.26^{***}$ \\
\quad Func.\ words forced deep       & $+0.02$       & $+0.06^{*}$ & $+0.14^{*}$ \\
\bottomrule
\end{tabular}
\caption{\textbf{Token-class gate overrides.} Per-token
cross-entropy increase relative to the learned gates, aggregated over
18 trained checkpoints spanning both FLOP budgets,
$\alpha \in \{25, 50, 75\}$, and $K \in \{2,3,4\}$. Arithmetic tokens
(\textsc{num}, \textsc{sym}) are forced wide ($g_d{=}0, g_w{=}1$);
function words (\textsc{adj}, \textsc{adv}, \textsc{part},
\textsc{pron}, \textsc{verb}) are forced deep ($g_d{=}1, g_w{=}0$);
each is compared against a size-matched random-token control under the
same override. The bottom block reports the paired per-checkpoint
excess of the class-targeted intervention over its control
(Wilcoxon signed-rank, $n{=}18$;
$^{*}p<0.05$, $^{***}p<0.001$).}
\label{tab:token-class-overrides}
\end{table*}

\begin{table*}[t]
\centering
\caption{Full results at the \textbf{F80M} FLOP budget. $\ell$ denotes the loop count. Commonsense Acc.\ / BPB are means over ARC-c, ARC-e, HellaSwag, PIQA, SIQA, WinoGrande. The OLMo3 easy math avg is the OLMo3 base-easy math BPB average. Best value in each row is \textbf{bold} (highest for accuracy $\uparrow$, lowest for BPB $\downarrow$).}
\label{tab:appendix-f80m}
\scriptsize
\setlength{\tabcolsep}{3pt}
\renewcommand{\arraystretch}{1.1}
\resizebox{\textwidth}{!}{%
\begin{tabular}{lccccccccccccc}
\toprule
\textbf{Metric} & \rotatebox[origin=lB]{60}{PureWide} & \rotatebox[origin=lB]{60}{PureLoop $\ell$2} & \rotatebox[origin=lB]{60}{PureLoop $\ell$3} & \rotatebox[origin=lB]{60}{PureLoop $\ell$4} & \rotatebox[origin=lB]{60}{Dual-a25 $\ell$2} & \rotatebox[origin=lB]{60}{Dual-a50 $\ell$2} & \rotatebox[origin=lB]{60}{Dual-a75 $\ell$2} & \rotatebox[origin=lB]{60}{Dual-a25 $\ell$3} & \rotatebox[origin=lB]{60}{Dual-a50 $\ell$3} & \rotatebox[origin=lB]{60}{Dual-a75 $\ell$3} & \rotatebox[origin=lB]{60}{Dual-a25 $\ell$4} & \rotatebox[origin=lB]{60}{Dual-a50 $\ell$4} & \rotatebox[origin=lB]{60}{Dual-a75 $\ell$4} \\
\textit{Params} & \textit{719M} & \textit{398M} & \textit{294M} & \textit{238M} & \textit{644M} & \textit{559M} & \textit{483M} & \textit{615M} & \textit{511M} & \textit{398M} & \textit{606M} & \textit{483M} & \textit{360M} \\
\midrule
\multicolumn{14}{l}{\textit{Commonsense (avg)}} \\
\quad Acc.\ $\uparrow$ & 0.5203 & 0.5036 & 0.5067 & 0.5031 & \textbf{0.5283} & 0.5265 & 0.5218 & 0.5239 & 0.5269 & 0.5192 & 0.5163 & 0.5256 & 0.5249 \\
\quad BPB $\downarrow$ & 0.9920 & 1.0028 & 1.0236 & 1.0352 & 0.9888 & 0.9826 & 0.9866 & 0.9827 & \textbf{0.9808} & 0.9870 & 0.9902 & 0.9820 & 0.9892 \\
\midrule
\multicolumn{14}{l}{\textit{Commonsense (per task, Acc.\ $\uparrow$ / BPB $\downarrow$)}} \\
\quad ARC-c Acc. & 0.4053 & 0.3779 & 0.3770 & 0.3809 & 0.4307 & 0.4111 & \textbf{0.4346} & 0.4336 & 0.4004 & 0.3994 & 0.4014 & 0.4170 & 0.4072 \\
\quad ARC-c BPB & 0.7967 & 0.8253 & 0.8546 & 0.8651 & 0.8001 & 0.7958 & \textbf{0.7954} & 0.7966 & 0.8016 & 0.8188 & 0.7974 & 0.7983 & 0.8090 \\
\quad ARC-e Acc. & 0.7158 & 0.6953 & 0.6973 & 0.6875 & 0.7148 & \textbf{0.7305} & 0.7100 & 0.6973 & 0.7178 & 0.7119 & 0.7148 & 0.6982 & 0.7266 \\
\quad ARC-e BPB & 0.5915 & 0.6134 & 0.6308 & 0.6801 & 0.5912 & \textbf{0.5779} & 0.5855 & 0.5907 & 0.5877 & 0.5820 & 0.5904 & 0.5912 & 0.6034 \\
\quad HellaSwag Acc. & 0.3906 & 0.3730 & 0.3760 & 0.3643 & 0.4014 & 0.4131 & 0.4043 & 0.4121 & \textbf{0.4150} & 0.3965 & 0.3955 & \textbf{0.4150} & 0.3955 \\
\quad HellaSwag BPB & 0.8947 & 0.8997 & 0.9156 & 0.9282 & 0.8872 & 0.8865 & 0.8928 & \textbf{0.8847} & 0.8888 & 0.8973 & 0.8940 & 0.8911 & 0.8946 \\
\quad PIQA Acc. & 0.6475 & 0.6348 & 0.6436 & 0.6338 & 0.6533 & 0.6436 & 0.6455 & 0.6348 & \textbf{0.6582} & 0.6357 & 0.6289 & 0.6455 & 0.6367 \\
\quad PIQA BPB & 1.1511 & 1.1816 & 1.1962 & 1.2069 & \textbf{1.1441} & 1.1524 & 1.1577 & 1.1504 & 1.1563 & 1.1646 & 1.1540 & 1.1627 & 1.1567 \\
\quad SIQA Acc. & 0.4453 & 0.4346 & 0.4297 & 0.4326 & \textbf{0.4492} & 0.4346 & 0.4346 & 0.4385 & 0.4443 & 0.4453 & 0.4424 & 0.4424 & \textbf{0.4492} \\
\quad SIQA BPB & 1.1863 & 1.2167 & 1.2252 & 1.2343 & 1.1983 & 1.2090 & 1.2101 & 1.1922 & \textbf{1.1625} & 1.1813 & 1.2147 & 1.1768 & 1.1975 \\
\quad WinoGrande Acc. & 0.5176 & 0.5059 & 0.5166 & 0.5195 & 0.5205 & 0.5264 & 0.5020 & 0.5273 & 0.5254 & 0.5264 & 0.5146 & \textbf{0.5352} & 0.5342 \\
\quad WinoGrande BPB & 1.3316 & 1.2803 & 1.3191 & 1.2964 & 1.3121 & 1.2741 & 1.2785 & 1.2819 & 1.2881 & 1.2782 & 1.2910 & \textbf{1.2716} & 1.2743 \\
\midrule
\multicolumn{14}{l}{\textit{Math}} \\
\quad GSM8k Acc.\ $\uparrow$ & 0.0791 & 0.0723 & 0.0723 & 0.0527 & 0.0879 & 0.0801 & \textbf{0.1006} & 0.0879 & \textbf{0.1006} & 0.0820 & 0.0664 & 0.0898 & 0.0918 \\
\quad Algebra BPB $\downarrow$ & 0.5201 & 0.5306 & 0.5385 & 0.5468 & 0.5173 & 0.5162 & 0.5198 & 0.5178 & \textbf{0.5115} & 0.5186 & 0.5215 & 0.5173 & 0.5141 \\
\quad Counting BPB $\downarrow$ & 0.5240 & 0.5343 & 0.5479 & 0.5557 & \textbf{0.5228} & 0.5244 & 0.5289 & 0.5259 & 0.5232 & 0.5294 & 0.5289 & 0.5230 & 0.5252 \\
\quad Geometry BPB $\downarrow$ & 0.5622 & 0.5736 & 0.5860 & 0.6004 & 0.5571 & 0.5579 & 0.5667 & 0.5596 & 0.5568 & 0.5634 & 0.5579 & \textbf{0.5567} & 0.5605 \\
\quad Int.\ Algebra BPB $\downarrow$ & 0.5402 & 0.5538 & 0.5592 & 0.5707 & 0.5332 & 0.5281 & 0.5409 & 0.5358 & 0.5317 & 0.5377 & 0.5372 & \textbf{0.5262} & 0.5334 \\
\quad Number Theory BPB $\downarrow$ & 0.5940 & 0.6057 & 0.6137 & 0.6218 & 0.5926 & 0.5902 & 0.5937 & 0.5924 & 0.5892 & 0.5981 & 0.5950 & \textbf{0.5870} & 0.5885 \\
\quad Pre-algebra BPB $\downarrow$ & 0.5069 & 0.5155 & 0.5237 & 0.5322 & 0.5061 & 0.5033 & 0.5087 & 0.5051 & \textbf{0.5005} & 0.5078 & 0.5095 & 0.5036 & 0.5048 \\
\quad Pre-calculus BPB $\downarrow$ & 0.4357 & 0.4470 & 0.4488 & 0.4670 & 0.4306 & 0.4252 & 0.4391 & 0.4316 & 0.4315 & 0.4335 & 0.4272 & \textbf{0.4247} & 0.4325 \\
\quad Easy math avg BPB $\downarrow$ & 0.5261 & 0.5372 & 0.5454 & 0.5564 & 0.5228 & 0.5208 & 0.5282 & 0.5240 & 0.5206 & 0.5269 & 0.5253 & \textbf{0.5198} & 0.5227 \\
\midrule
\multicolumn{14}{l}{\textit{Reading / QA}} \\
\quad LAMBADA Acc.\ $\uparrow$ & 0.2861 & 0.2803 & 0.2744 & 0.2793 & 0.2949 & 0.2725 & 0.2939 & 0.2832 & \textbf{0.3135} & 0.2881 & 0.2988 & 0.3037 & 0.3076 \\
\quad LAMBADA BPB $\downarrow$ & 0.9492 & 0.9769 & 0.9887 & 1.0032 & 0.9161 & 0.9454 & 0.9342 & 0.9167 & \textbf{0.9146} & 0.9462 & 0.9341 & 0.9154 & 0.9473 \\
\quad QASPER Acc.\ $\uparrow$ & 0.5987 & 0.6740 & 0.6771 & 0.6771 & 0.6646 & 0.5361 & 0.6677 & 0.6301 & 0.5862 & \textbf{0.6803} & 0.6301 & 0.4765 & 0.6771 \\
\quad QASPER BPB $\downarrow$ & 0.3063 & 0.3098 & 0.3150 & 0.3233 & 0.3012 & 0.3030 & 0.3010 & 0.3035 & 0.3027 & 0.3020 & \textbf{0.2992} & 0.3108 & 0.3081 \\
\midrule
\multicolumn{14}{l}{\textit{Paloma (BPB $\downarrow$)}} \\
\quad C4 & 1.0370 & 1.0513 & 1.0652 & 1.0777 & \textbf{1.0311} & 1.0374 & 1.0395 & 1.0331 & 1.0358 & 1.0472 & 1.0374 & 1.0362 & 1.0451 \\
\quad WikiText-103 & 0.9461 & 0.9608 & 0.9757 & 0.9918 & 0.9373 & 0.9453 & 0.9477 & \textbf{0.9344} & 0.9382 & 0.9508 & 0.9420 & 0.9393 & 0.9480 \\
\bottomrule
\end{tabular}%
}
\end{table*}

\begin{table*}[t]
\centering
\caption{Full results at the \textbf{F160M} FLOP budget. $\ell$ denotes the loop count. Commonsense Acc.\ / BPB are means over ARC-c, ARC-e, HellaSwag, PIQA, SIQA, WinoGrande. The OLMo3 easy math avg is the OLMo3 base-easy math BPB average. Best value in each row is \textbf{bold} (highest for accuracy $\uparrow$, lowest for BPB $\downarrow$).}
\label{tab:appendix-f160m}
\scriptsize
\setlength{\tabcolsep}{3pt}
\renewcommand{\arraystretch}{1.1}
\resizebox{\textwidth}{!}{%
\begin{tabular}{lccccccccccccc}
\toprule
\textbf{Metric} & \rotatebox[origin=lB]{60}{PureWide} & \rotatebox[origin=lB]{60}{PureLoop $\ell$2} & \rotatebox[origin=lB]{60}{PureLoop $\ell$3} & \rotatebox[origin=lB]{60}{PureLoop $\ell$4} & \rotatebox[origin=lB]{60}{Dual-a25 $\ell$2} & \rotatebox[origin=lB]{60}{Dual-a50 $\ell$2} & \rotatebox[origin=lB]{60}{Dual-a75 $\ell$2} & \rotatebox[origin=lB]{60}{Dual-a25 $\ell$3} & \rotatebox[origin=lB]{60}{Dual-a50 $\ell$3} & \rotatebox[origin=lB]{60}{Dual-a75 $\ell$3} & \rotatebox[origin=lB]{60}{Dual-a25 $\ell$4} & \rotatebox[origin=lB]{60}{Dual-a50 $\ell$4} & \rotatebox[origin=lB]{60}{Dual-a75 $\ell$4} \\
\textit{Params} & \textit{1361M} & \textit{719M} & \textit{502M} & \textit{398M} & \textit{1200M} & \textit{1040M} & \textit{880M} & \textit{1153M} & \textit{936M} & \textit{719M} & \textit{1125M} & \textit{880M} & \textit{644M} \\
\midrule
\multicolumn{14}{l}{\textit{Commonsense (avg)}} \\
\quad Acc.\ $\uparrow$ & 0.5384 & 0.5303 & 0.5244 & 0.5137 & 0.5374 & 0.5358 & 0.5202 & \textbf{0.5493} & 0.5360 & 0.5384 & 0.5420 & 0.5326 & 0.5335 \\
\quad BPB $\downarrow$ & 0.9636 & 0.9814 & 0.9894 & 1.0010 & 0.9755 & 0.9608 & 0.9744 & 0.9655 & 0.9572 & \textbf{0.9572} & 0.9610 & 0.9704 & 0.9700 \\
\midrule
\multicolumn{14}{l}{\textit{Commonsense (per task, Acc.\ $\uparrow$ / BPB $\downarrow$)}} \\
\quad ARC-c Acc. & 0.4277 & 0.4170 & 0.4053 & 0.3896 & 0.4365 & 0.4160 & 0.4150 & 0.4346 & 0.4326 & 0.4316 & 0.4482 & \textbf{0.4512} & 0.4375 \\
\quad ARC-c BPB & 0.7715 & 0.7880 & 0.8107 & 0.8178 & 0.7758 & 0.7674 & 0.7746 & 0.7839 & \textbf{0.7652} & 0.7704 & 0.7800 & 0.7673 & 0.7730 \\
\quad ARC-e Acc. & \textbf{0.7520} & 0.7266 & 0.7109 & 0.7051 & 0.7207 & 0.7275 & 0.7129 & 0.7461 & 0.7305 & 0.7314 & 0.7373 & 0.7324 & 0.7363 \\
\quad ARC-e BPB & \textbf{0.5558} & 0.5717 & 0.5961 & 0.6046 & 0.5634 & 0.5579 & 0.5682 & 0.5610 & 0.5633 & 0.5606 & 0.5624 & 0.5775 & 0.5668 \\
\quad HellaSwag Acc. & 0.4229 & 0.4248 & 0.4004 & 0.4043 & \textbf{0.4414} & 0.4316 & 0.4111 & 0.4404 & 0.4258 & 0.4219 & 0.4277 & 0.4209 & 0.4170 \\
\quad HellaSwag BPB & 0.8767 & 0.8809 & 0.8935 & 0.8929 & 0.8670 & 0.8723 & 0.8827 & 0.8696 & 0.8705 & 0.8738 & \textbf{0.8661} & 0.8796 & 0.8759 \\
\quad PIQA Acc. & 0.6689 & 0.6445 & 0.6455 & 0.6367 & \textbf{0.6777} & 0.6592 & 0.6436 & 0.6660 & 0.6543 & 0.6611 & 0.6494 & 0.6602 & 0.6562 \\
\quad PIQA BPB & 1.1263 & 1.1494 & 1.1660 & 1.1705 & \textbf{1.1145} & 1.1259 & 1.1336 & 1.1181 & 1.1185 & 1.1304 & 1.1367 & 1.1354 & 1.1349 \\
\quad SIQA Acc. & 0.4541 & 0.4463 & 0.4414 & 0.4395 & 0.4395 & 0.4541 & 0.4463 & \textbf{0.4658} & 0.4434 & 0.4492 & 0.4502 & 0.4277 & 0.4443 \\
\quad SIQA BPB & 1.1906 & 1.2077 & 1.2048 & 1.2272 & \textbf{1.1524} & 1.1695 & 1.1809 & 1.1780 & 1.1751 & 1.1639 & 1.1638 & 1.1794 & 1.1893 \\
\quad WinoGrande Acc. & 0.5049 & 0.5225 & \textbf{0.5430} & 0.5068 & 0.5088 & 0.5264 & 0.4922 & \textbf{0.5430} & 0.5293 & 0.5352 & 0.5391 & 0.5029 & 0.5098 \\
\quad WinoGrande BPB & 1.2609 & 1.2908 & 1.2650 & 1.2932 & 1.3800 & 1.2719 & 1.3063 & 1.2826 & 1.2509 & \textbf{1.2439} & 1.2572 & 1.2830 & 1.2802 \\
\midrule
\multicolumn{14}{l}{\textit{Math}} \\
\quad GSM8k Acc.\ $\uparrow$ & 0.1250 & 0.1045 & 0.1191 & 0.0938 & 0.0996 & 0.1084 & 0.1055 & 0.1084 & 0.1074 & 0.1094 & 0.1250 & 0.1328 & \textbf{0.1406} \\
\quad Algebra BPB $\downarrow$ & 0.5038 & 0.5013 & 0.5101 & 0.5114 & 0.5074 & 0.5081 & 0.5049 & 0.5043 & 0.5007 & 0.5041 & 0.5018 & \textbf{0.4920} & 0.4976 \\
\quad Counting BPB $\downarrow$ & 0.5115 & 0.5112 & 0.5200 & 0.5242 & 0.5127 & 0.5147 & 0.5157 & 0.5135 & 0.5098 & 0.5129 & 0.5129 & \textbf{0.5031} & 0.5083 \\
\quad Geometry BPB $\downarrow$ & 0.5419 & 0.5426 & 0.5466 & 0.5585 & 0.5397 & 0.5457 & 0.5478 & 0.5383 & 0.5389 & 0.5440 & 0.5369 & \textbf{0.5282} & 0.5404 \\
\quad Int.\ Algebra BPB $\downarrow$ & 0.5137 & 0.5201 & 0.5250 & 0.5282 & 0.5208 & 0.5215 & 0.5243 & 0.5175 & 0.5186 & 0.5191 & 0.5154 & \textbf{0.5003} & 0.5145 \\
\quad Number Theory BPB $\downarrow$ & 0.5751 & 0.5783 & 0.5840 & 0.5872 & 0.5804 & 0.5840 & 0.5784 & 0.5804 & 0.5769 & 0.5758 & 0.5758 & \textbf{0.5632} & 0.5699 \\
\quad Pre-algebra BPB $\downarrow$ & 0.4923 & 0.4924 & 0.4974 & 0.5007 & 0.4942 & 0.4964 & 0.4956 & 0.4920 & 0.4913 & 0.4928 & 0.4917 & \textbf{0.4831} & 0.4864 \\
\quad Pre-calculus BPB $\downarrow$ & 0.4154 & 0.4177 & 0.4221 & 0.4289 & 0.4210 & 0.4231 & 0.4239 & 0.4135 & 0.4137 & 0.4165 & 0.4117 & \textbf{0.4010} & 0.4155 \\
\quad Easy math avg BPB $\downarrow$ & 0.5077 & 0.5091 & 0.5150 & 0.5199 & 0.5109 & 0.5134 & 0.5130 & 0.5085 & 0.5071 & 0.5093 & 0.5066 & \textbf{0.4959} & 0.5047 \\
\midrule
\multicolumn{14}{l}{\textit{Reading / QA}} \\
\quad LAMBADA Acc.\ $\uparrow$ & 0.3086 & 0.3135 & 0.2920 & 0.2939 & 0.3115 & 0.2979 & 0.2822 & 0.2959 & 0.3115 & 0.3223 & 0.3105 & \textbf{0.3271} & 0.3203 \\
\quad LAMBADA BPB $\downarrow$ & 0.9169 & 0.9038 & 0.9408 & 0.9359 & 0.8811 & 0.8997 & 0.9184 & 0.9031 & 0.8985 & 0.8780 & 0.8854 & \textbf{0.8646} & 0.8945 \\
\quad QASPER Acc.\ $\uparrow$ & 0.5768 & 0.6301 & 0.6583 & 0.5987 & 0.6583 & 0.6207 & 0.6803 & \textbf{0.6865} & 0.5799 & 0.6301 & 0.4514 & 0.5925 & 0.6050 \\
\quad QASPER BPB $\downarrow$ & 0.2964 & 0.2990 & \textbf{0.2916} & 0.2941 & 0.3120 & 0.3405 & 0.2982 & 0.3112 & 0.3215 & 0.2963 & 0.3247 & 0.3309 & 0.2934 \\
\midrule
\multicolumn{14}{l}{\textit{Paloma (BPB $\downarrow$)}} \\
\quad C4 & 1.0183 & 1.0270 & 1.0367 & 1.0446 & 1.0130 & 1.0163 & 1.0203 & 1.0138 & 1.0138 & 1.0188 & \textbf{1.0129} & 1.0166 & 1.0212 \\
\quad WikiText-103 & 0.9224 & 0.9312 & 0.9388 & 0.9453 & 0.9137 & 0.9208 & 0.9220 & 0.9177 & 0.9147 & 0.9214 & \textbf{0.9108} & 0.9201 & 0.9241 \\
\bottomrule
\end{tabular}%
}
\end{table*}

\begin{table*}[t]
\centering
\small
\setlength{\tabcolsep}{6pt}
\begin{tabular}{l r r r}
\toprule
\textbf{Ablation} & \textbf{GSM8K} & \textbf{TriviaQA} & \textbf{WikiText-103} \\
\midrule
Learned router (baseline)                  & 2.148 & 4.129 & 3.026 \\
\midrule
\multicolumn{4}{l}{\emph{Compute (force $K$ loops; trained with $K{=}4$)}} \\
\quad $K{=}1$                              & 3.592 \ssub{+1.44}   & 5.777 \ssub{+1.65}   & 4.846 \ssub{+1.82}   \\
\quad $K{=}2$                              & 2.665 \ssub{+0.52}   & 4.694 \ssub{+0.57}   & 3.842 \ssub{+0.82}   \\
\quad $K{=}3$                              & 2.221 \ssub{+0.07}   & 4.174 \ssub{+0.05}   & 3.178 \ssub{+0.15}   \\
\quad $K{=}4$ (\emph{trained})             & 2.148 \ssub{+0.00}   & 4.129 \ssub{+0.00}   & 3.026 \ssub{+0.00}   \\
\quad $K{=}5$                              & 2.206 \ssub{+0.06}   & 4.187 \ssub{+0.06}   & 3.046 \ssub{+0.02}   \\
\quad $K{=}6$                              & 2.271 \ssub{+0.12}   & 4.239 \ssub{+0.11}   & 3.074 \ssub{+0.05}   \\
\quad $K{=}7$                              & 2.323 \ssub{+0.18}   & 4.283 \ssub{+0.15}   & 3.098 \ssub{+0.07}   \\
\quad $K{=}8$                              & 2.366 \ssub{+0.22}   & 4.320 \ssub{+0.19}   & 3.120 \ssub{+0.09}   \\
\quad $K{=}9$                              & 2.405 \ssub{+0.26}   & 4.351 \ssub{+0.22}   & 3.140 \ssub{+0.11}   \\
\quad $K{=}10$                             & 2.442 \ssub{+0.29}   & 4.380 \ssub{+0.25}   & 3.160 \ssub{+0.13}   \\
\midrule
\multicolumn{4}{l}{\emph{Gate overrides}} \\
\quad $g_d{=}1,\ g_w{=}0$ (deep only)      & 7.684 \ssub{+5.54}   & 8.291 \ssub{+4.16}   & 8.090 \ssub{+5.06}   \\
\quad $g_d{=}0,\ g_w{=}1$ (wide only)      & 7.574 \ssub{+5.43}   & 7.747 \ssub{+3.62}   & 8.655 \ssub{+5.63}   \\
\quad $g_d{=}0.5,\ g_w{=}0.5$ (uniform)    & 4.744 \ssub{+2.60}   & 6.707 \ssub{+2.58}   & 5.749 \ssub{+2.72}   \\
\quad $g_d{=}1,\ g_w{=}1$ (fully open)     & 8.541 \ssub{+6.39}   & 9.099 \ssub{+4.97}   & 9.211 \ssub{+6.19}   \\
\quad Shuffled gates                       & 2.658 \ssub{+0.51}   & 5.071 \ssub{+0.94}   & 3.720 \ssub{+0.69}   \\
\bottomrule
\end{tabular}
\caption{%
Inference-time ablations of the dual-path router. We report per-token
cross-entropy loss on three Paloma sources. \textbf{Compute (top block):}
forcing exactly $K$ loop iterations confirms that loop budget matters,
with diminishing returns past the trained value. Extra inference loops beyond training degrade
the model monotonically rather than plateauing, indicating the loop schedule does not extrapolate past its training budget. \textbf{Gate overrides (bottom block):} both paths are necessary, disabling either ($g_d{=}1,g_w{=}0$ or $g_d{=}0,g_w{=}1$) collapses
performance. 
}
\label{tab:ablations}
\end{table*}

\begin{table}[h]
\centering
\small
\setlength{\tabcolsep}{4pt}
\renewcommand{\arraystretch}{1.05}
\begin{tabular}{lc cccc cccc}
\toprule
 & & \multicolumn{4}{c}{$F_M = 80$M} & \multicolumn{4}{c}{$F_M = 160$M} \\
\cmidrule(lr){3-6}\cmidrule(lr){7-10}
Config & $K$ & $d_{\text{ffn}}$ & $d_{\text{ffn}}^{\text{wide}}$ & Params & Time & $d_{\text{ffn}}$ & $d_{\text{ffn}}^{\text{wide}}$ & Params & Time \\
\midrule
\textsc{PureWide}                    & 1 & ---   & 24576 & 719M & 17.6h & ---   & 50624 & 1361M & 21.4h \\
\textsc{PureLoop} $K{=}2$            & 2 & 11392 & ---   & 398M & 15.2h & 24448 & ---   &  719M & 19.9h \\
\textsc{PureLoop} $K{=}3$            & 3 &  7104 & ---   & 294M & 16.6h & 15744 & ---   &  502M & 15.5h \\
\textsc{PureLoop} $K{=}4$            & 4 &  4864 & ---   & 238M & 15.9h & 11392 & ---   &  398M & 17.7h \\
\midrule
\textsc{Dual} $\alpha{=}25$, $K{=}2$ & 2 &  1600 & 17920 & 644M & 14.6h &  4864 & 37440 & 1200M & 14.8h \\
\textsc{Dual} $\alpha{=}25$, $K{=}3$ & 3 &   576 & 17920 & 615M & 14.7h &  2752 & 37440 & 1153M & 15.1h \\
\textsc{Dual} $\alpha{=}25$, $K{=}4$ & 4 &    64 & 17920 & 606M & 12.8h &  1600 & 37440 & 1125M & 15.0h \\
\textsc{Dual} $\alpha{=}50$, $K{=}2$ & 2 &  4864 & 11392 & 559M & 17.4h & 11392 & 24448 & 1040M & 21.3h \\
\textsc{Dual} $\alpha{=}50$, $K{=}3$ & 3 &  2752 & 11392 & 511M & 17.0h &  7104 & 24448 &  936M & 15.6h \\
\textsc{Dual} $\alpha{=}50$, $K{=}4$ & 4 &  1600 & 11392 & 483M & 16.5h &  4864 & 24448 &  880M & 19.0h \\
\textsc{Dual} $\alpha{=}75$, $K{=}2$ & 2 &  8128 &  4864 & 483M & 14.2h & 17920 & 11392 &  880M & 13.2h \\
\textsc{Dual} $\alpha{=}75$, $K{=}3$ & 3 &  4864 &  4864 & 398M & 13.3h & 11392 & 11392 &  719M & 14.5h \\
\textsc{Dual} $\alpha{=}75$, $K{=}4$ & 4 &  3264 &  4864 & 360M & 14.3h &  8128 & 11392 &  644M & 13.1h \\
\bottomrule
\end{tabular}%
\caption{Per-configuration widths, parameter counts, and wall-clock
training times at the $F_M = 80$M and $F_M = 160$M FFN-FLOP budgets.}
\label{tab:configs-f80m}
\end{table}

\end{document}